\documentclass[10pt]{article} 
\usepackage[accepted]{tmlr}

\usepackage{amsmath,amsfonts,bm}

\def\eqref#1{equation~\ref{#1}}

\def\1{\bm{1}}

\DeclareMathAlphabet{\mathsfit}{\encodingdefault}{\sfdefault}{m}{sl}
\SetMathAlphabet{\mathsfit}{bold}{\encodingdefault}{\sfdefault}{bx}{n}

\usepackage{hyperref}
\usepackage{url}
\usepackage{graphicx}
\usepackage{subfigure}
\usepackage{threeparttable}
\usepackage{multirow}
\usepackage{comment}
\usepackage{enumitem}
\usepackage[utf8]{inputenc} 
\usepackage[T1]{fontenc}    
\usepackage{url}            
\usepackage{booktabs}       
\usepackage{amsfonts}       
\usepackage{nicefrac}       
\usepackage{microtype}      
\usepackage{subcaption}
\usepackage{makecell}
\usepackage{wrapfig,lipsum,booktabs}

\newcommand\rebuttal[1]{\textcolor{black}{#1}}

\newcommand{\address}[1]{\textit{#1}}

\title{ReDistill: Residual Encoded Distillation for \\Peak Memory Reduction \rebuttal{of CNNs}}

\author{\name Fang Chen$^1$ \email fchen20@ucmerced.edu 
      \AND
      \name Gourav Datta$^2$ \email  gourav.datta@case.edu 
      \AND
      \name Mujahid Al Rafi$^1$ \email mrafi@ucmerced.edu 
      \AND
      \name Hyeran Jeon$^1$ \email 
      hjeon7@ucmerced.edu 
      \AND
      \name Meng Tang$^1$ \email mtang4@ucmerced.edu\\
      \\
      \address{$^{1}$University of California Merced} \\ 
      \address{$^{2}$Case Western Reserve University}
}

\def\month{04}  
\def\year{2025} 
\def\openreview{\url{https://openreview.net/forum?id=akumIxQjNN}} 

\begin{document}

\maketitle

\begin{abstract}
 The expansion of neural network sizes and the enhanced resolution of modern image sensors result in heightened memory and power demands to process modern computer vision models. 
 In order to deploy these models in extremely resource-constrained edge devices, it is crucial to reduce their peak memory, which is the maximum memory consumed during the execution of a model.
 A naive approach to reducing peak memory is aggressive down-sampling of feature maps via pooling with large stride, which often results in unacceptable degradation in network performance.
 To mitigate this problem, we propose residual encoded distillation (ReDistill) for peak memory reduction in a teacher-student framework, in which a student network with less memory is derived from the teacher network using aggressive pooling.
 We apply our distillation method to multiple problems in computer vision, including image classification and diffusion-based image generation.
 For image classification, our method yields $\textbf{4x-5x}$ theoretical peak memory reduction with less degradation in accuracy for most CNN-based architectures. 
 For diffusion-based image generation, our proposed distillation method yields a denoising network with $\textbf{4x}$ lower theoretical peak memory while maintaining decent diversity and fidelity for image generation.
 Experiments demonstrate our method's superior performance compared to other feature-based and response-based distillation methods when applied to the same student network.
 The code is available at \href{https://github.com/mengtang-lab/ReDistill}{https://github.com/mengtang-lab/ReDistill}.
\end{abstract}

\section{Introduction}
\label{introduction}

Convolutional neural networks (CNN) have demonstrated impressive capabilities across diverse computer vision tasks such as image recognition~(\cite{simonyan2014very}), object detection~(\cite{redmon2018yolov3}), semantic segmentation~(\cite{long2015fully}), and image generation~(\cite{creswell2018generative}).
However, the ever-growing network size and image resolution of modern imaging sensors pose significant challenges in deploying neural networks on standard edge devices with limited memory footprint.
For example, a standard STM32H5 MCU provides only 640 KB of SRAM and 2 MB of Flash storage. These constraints make it impractical to execute off-the-shelf deep learning models: ResNet-50 surpasses the storage limit by $44\times$, while MobileNetV2 exceeds the peak memory limit by $8\times$. Even the int8 quantized version of MobileNetV2 surpasses the memory limit by $2\times$, underscoring a substantial disparity between desired and available hardware capacity.
Hence, it is very important to reduce the \textit{peak memory} during inference 
for edge deployment. Note that our primary focus in this work is reducing peak memory usage, as there are existing solutions for addressing other metrics, such as parameter count and the number of operations when deploying CV models at the extreme edge.
Similar to~(\cite{mcunetv2,chowdhery2019visual}), we estimate the theoretical peak memory by summing the size of the input \& output allocation for each operation (e.g., convolution, non-linear activation, pooling).
%
Through empirical measurements, we determined that peak memory usage is predominantly influenced by the initial layers of convolutional neural networks (CNNs) that are characterized by large feature maps. For U-shaped CNN architectures, however, the last few layers also significantly contribute to peak memory consumption.
%


\begin{figure}
\centerline{
\includegraphics[scale=0.36]{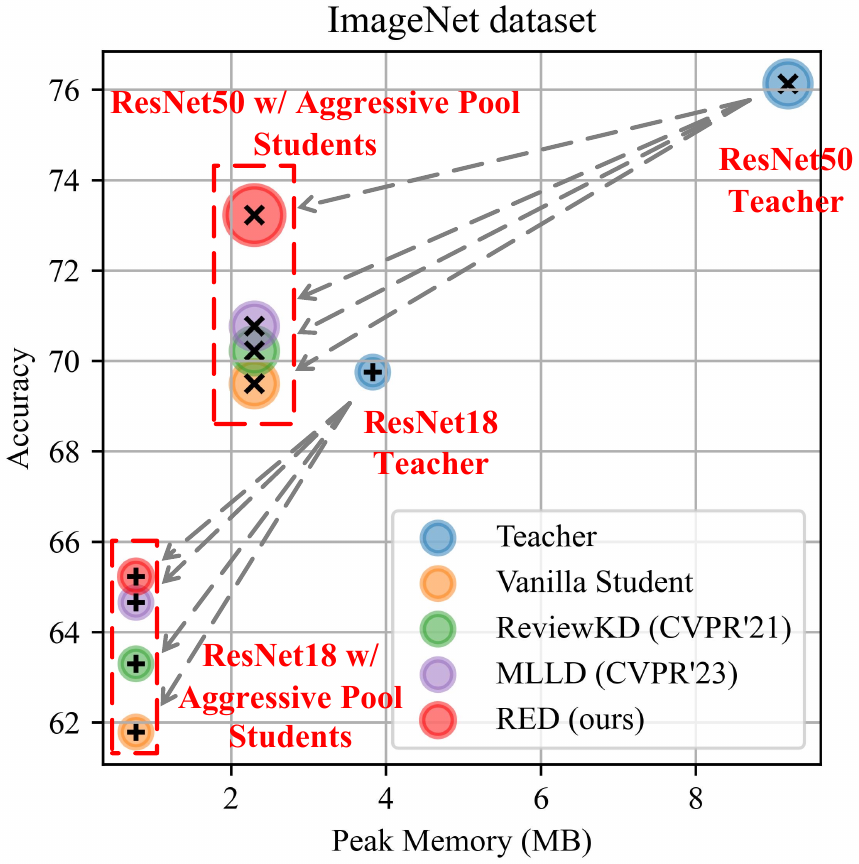}
\includegraphics[scale=0.3]{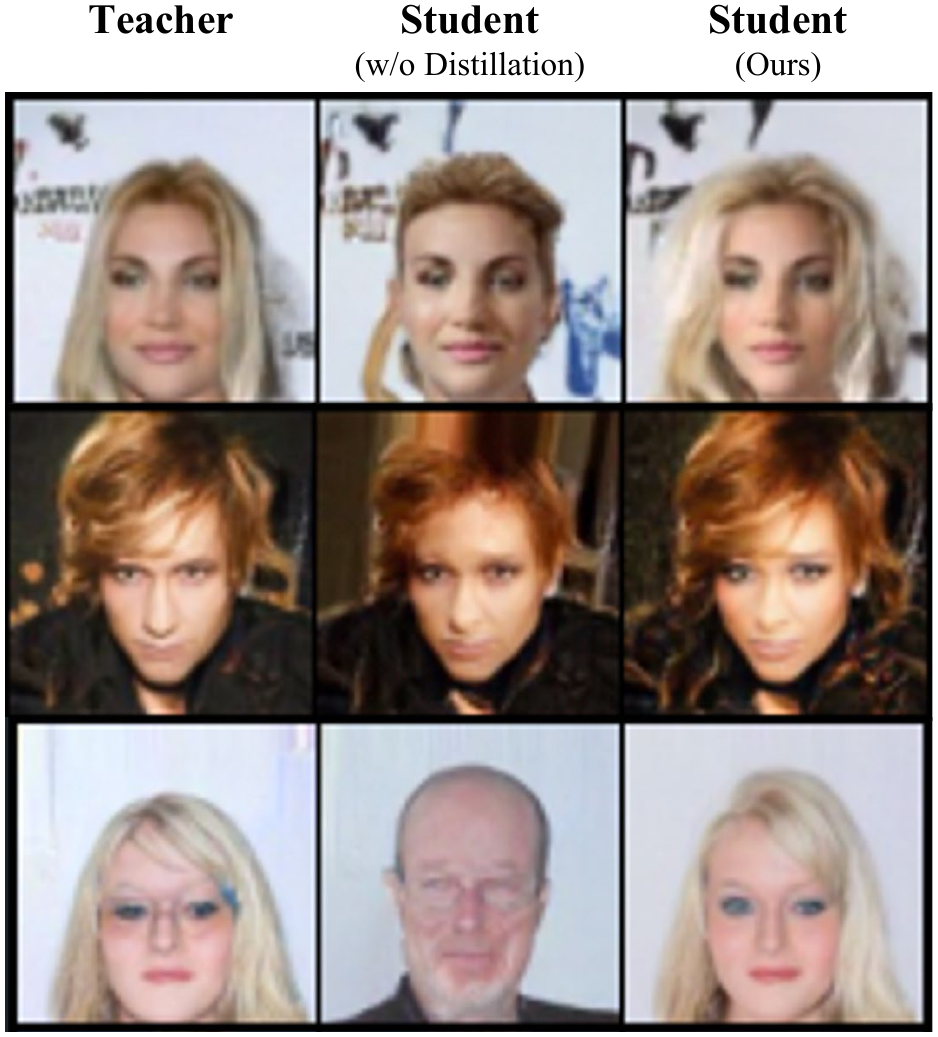}
}
\caption{(a) \textit{Left}: For ImageNet classification, our distillation method significantly reduces the theoretical peak memory of ResNet-based models while achieving accuracy better than existing distillation methods. 
(b) \textit{Right}: For diffusion-based image generation, our distilled network with $4{\times}$ lower theoretical peak memory generates images indistinguishable from the generated images of a teacher network.}
\label{fig:teaser}
\end{figure}

A naive approach to reducing peak memory is aggressive downsampling via pooling with large kernel size and large stride, which often leads to unacceptable degradation of network performance due to loss of information in small feature maps.
Given a teacher network with large peak memory, we propose residual encoded distillation (ReDistill) to train a student network with significantly lower peak memory. This student network can be considered a variant of the teacher network with aggressive pooling. %
We demonstrate the effectiveness of our methods for multiple problems including image classification and diffusion-based image synthesis.
For image classification with ResNet-based models shown in Fig.~\ref{fig:teaser}~(a), our method reduces the theoretical peak memory by $4{-}5\times$ with a lower accuracy drop compared to existing distillation methods.
%
For diffusion-based image generation shown in Fig.~\ref{fig:teaser}~(b), our distilled network generates images similar to original networks, yet the theoretical peak memory is reduced by $4\times$ on average.

Our ReDistill method outperforms existing response-based or feature-based distillation methods regarding the accuracy-memory trade-off.
Our method differs from these existing counterparts in four regards.
Firstly, our distillation method is tailored for peak memory reduction.
In contrast, existing distillation techniques focus on transferring knowledge from a high-capability teacher network with a large number of parameters to a student network with fewer parameters.
Our student networks apply a large kernel size and stride in the initial pooling layers with the same number of parameters as the teacher networks while consuming significantly lower peak memory.
Secondly, the student network, utilizing aggressive pooling, has fewer pooling layers and consequently fewer stages than the teacher network, resulting in mismatched features at different stages between the two.
We add novel non-linear mapping modules termed residual encoded distillation (RED) blocks between the teacher and student network during both training and inference. 
Thirdly, our proposed RED block 
is lightweight and effective with additive residual learning and multiplicative gating mechanism.
We optimize the trade-off between peak memory and accuracy, while it slightly increases the model size due to extra parameters.
Lastly, we align teacher and student network features asynchronously at pooling layers with matching feature sizes, while previous approaches align features at different stages of the networks.

Our key contributions are summarized below.

\begin{itemize}[leftmargin=*]
\item We propose ReDistill, a distillation framework tailored for reducing the peak memory of convolutional neural networks.
    Our method allows aggressive downsampling of feature maps via pooling layers with a large stride for a student network while incurring a less accuracy drop.
    To the best of our knowledge, ReDistill is the first distillation method focused on peak memory reduction for efficient deep learning.
    \item The core of our ReDistill framework is a residual encoded distillation (RED) block to align features between high-peak-memory teacher networks and low-peak-memory student networks.
    Our RED block is based on a multiplicative gating mechanism and additive residual learning and is shown to be simple and effective for peak memory reduction with minimum computational overhead.
    \item For image classification tasks, our distillation method outperforms state-of-the-art response-based 
    or feature-based distillation methods when applied to the same student network assigned with a large pooling stride, as shown in extensive experiments with multiple datasets.
    Our method yields $4\times\sim5\times$ reduction in theoretical peak memory with a slight decrease in classification accuracies for CNN based models. 
    \item We also show the versatility of our distillation method for denoising diffusion probabilistic models for image generation.
    For a U-Net based denoising network, our method reduces the theoretical peak memory by $4{\times}$ by downsampling the feature maps of the first few encoder layers and last few decoder layers while maintaining the fidelity and diversity of synthesized images.
\end{itemize}


\section{Related Work}
\label{related work}

\paragraph{Memory-constrained deep learning} Limited memory capacity in GPU cards and edge platforms has been a critical hurdle in CNN training and inference.
%
%
Multiple GPUs can be utilized through model and data parallelism~(\cite{langer2020distributed}) to mitigate the memory bottleneck.
Other solutions include optimization methods such as network quantization~(\cite{hubara2016binarized}), compression~(\cite{han2016deep}), and pruning~(\cite{molchanov2017pruning,lu2024entropy}), which focus on maintaining essential bits of weights or parameters while minimizing accuracy loss.
To produce correct outputs with compressed data, these solutions are typically designed with specialized accelerators to accommodate meta-data processing~(\cite{han2016eie}).
There are also CNNs specifically designed for resource-constrained applications, such as variants of MobileNet~(\cite{mobilenetv3}) and SqueezeNet~(\cite{iandola2017squeezenet}).
%
%
These approaches to memory-constrained deep learning are orthogonal to our ReDistill framework, which focuses on peak memory reduction during inference. 
Nevertheless, recent work has explored neural architecture search (NAS) to create networks with minimized peak memory~(\cite{mcunet}).
Reference~(\cite{mcunetv2}) takes this a step further by leveraging NAS to introduce patch-based inference and network redistribution~(\cite{mcunetv2}), consequently shifting the receptive field to later stages. While NAS significantly exacerbates the training complexity, patch-based inference necessitates compiler libraries that may not be compatible with standard GPUs and incurs additional computation and latency overhead. Another recent work~(\cite{chen2023self}) proposed self-attention-based pooling to aggressively compress the activation maps in the first few layers to reduce the peak memory at the cost of increased compute complexity.
%
\paragraph{Knowledge distillation for image classification} can be roughly categorized into two groups: response-based KD and feature-based KD.
 Response-based KD methods derive the distillation loss by leveraging the logit outputs from the fully connected layers of the student model and the teacher model.
 For example, KD~(\cite{KD2015}) distills knowledge by matching the prediction probability distributions of the student architecture and the teacher architecture.
 DKD~(\cite{DKD2022}) decouples the classical KD loss into two parts, target class knowledge distillation (TCKD) and non-target class knowledge distillation (NCKD) enhancing training efficiency and flexibility.
 MLLD~(\cite{MLLD2023}) performs logit distillation through a multi-level alignment based on instance prediction, input correlation, and category correlation, delivering state-of-the-art performance.
 In contrast, feature-based KD methods~(\cite{FITNET2015, AT2016, PKT2018, KDSVD2018, Similarity2019, VID2019}) reduce the disparity between features in the teacher and student models, compelling the student model to replicate the teacher model at the feature level.
 RKD~(\cite{RKD2019}) employs a relation potential function to convey information from the teacher's features to the student's features.
 ReviewKD~(\cite{ReviewKD2021}) aggregates knowledge of the teacher from different stages into one stage of the student, the so-called `knowledge review', which achieved impressive performance.
 KCD~(\cite{KCD2022}) iteratively condenses a compact knowledge set from the teacher to guide the student learning by the Expectation-Maximization (EM) algorithm, which would empower and be easily applied to other knowledge distillation algorithms.
 Existing methods focus on the distillation from a high-capacity teacher model with a large amount of parameters to an efficient student model with limited parameters.
 In this work, the student model, employing a large kernel size and stride in the initial pooling layer 
 possesses the same number of parameters as the teacher architecture but incurs significantly lower peak memory. \\
\paragraph{Knowledge distillation for diffusion models} has gained popularity.
For example, One Step Diffusion~(\cite{yin2023one}) defines two score functions, one of the target distribution and the other of the synthetic distribution produced by a one-step generator.
 By minimizing the KL divergence between these two score functions, the one-step generator is enforced to match the diffusion model at the distribution level and achieves impressive performance.
 Adversarial Diffusion~(\cite{sauer2023adversarial}) utilizes both score distillation loss and adversarial loss. 
 The score distillation loss occurs between the teacher diffusion sampler with a large number of $T$ steps and the student diffusion sampler with one or two steps.
 Meanwhile, the adversarial loss originates from a discriminator trained to differentiate between generated samples and real images.
 Auto Diffusion~(\cite{li2023autodiffusion}) searches for the optimal time steps and compressed models in a unified framework to achieve effective image generation for diffusion models.
 In summary, existing KD methods for diffusion models mainly focus on time-step reduction and model compression.
 We offer a unique and orthogonal approach by minimizing peak memory that can be easily integrated with existing methods.

\vspace{-2mm}

\section{Proposed Method}
\label{proposed method}

\subsection{Preliminaries}

 \textbf{Knowledge Distillation}\quad We are given a dataset $\mathcal{X}$, a high-capacity teacher architecture $\mathcal{T}$ and a to-learned student architecture $\mathcal{S}$.
 For an input image $x$ sampled from $\mathcal{X}$, $\pi_\mathcal{T}(x)$ and $\pi_\mathcal{S}(x)$ denote the outputs or intermediate features of the teacher and student, respectively.
 The knowledge distillation task aims to optimize the student's parameters $\hat{w}$:
 \begin{small}
 \begin{eqnarray}\label{eq:knowledge distillation}
 \hat{w} = \mathop{\arg\min}\limits_{w}\sum_{x\in\mathcal{X}}{\mathcal{L}(\pi_\mathcal{S}(x;w), \pi_\mathcal{T}(x))},
 \end{eqnarray}
 \end{small}where $w$ denotes the trainable weights of $\pi_\mathcal{S}$ and $\mathcal{L}$ denotes the loss function defined by different knowledge distillation methods.
 For instance, \cite{KD2015} defines $\pi_\mathcal{S}$ and $\pi_\mathcal{T}$ as the logit outputs (without applying the softmax function) of the student and the teacher, while $\mathcal{L}$ as the Kullback-Leibler divergence between $\pi_\mathcal{S}$ and $\pi_\mathcal{T}$ after applying the softmax function with temperature $t_p$:
 \begin{small}
 \begin{eqnarray}\label{eq:hinton's kd}
 \mathcal{L}_{KL} = \quad\sum_{x\in\mathcal{X}}{KL(softmax(\frac{\pi_\mathcal{S}(x)}{t_p}), softmax(\frac{\pi_\mathcal{T}(x))}{t_p})}.
 \end{eqnarray}
 \end{small}Some other methods~(\cite{AT2016, RKD2019, ReviewKD2021}) define different $\pi_\mathcal{S}$ and $\pi_\mathcal{T}$, such as the intermediate activation maps from various stages of the student and teacher, or different $\mathcal{L}$, like the $p$-norm, to achieve various distillation methods.
 
\rebuttal{
 KD methods can be categorized as response-based methods with $\pi_\mathcal{S}, \pi_\mathcal{T}$ defined as the logit outputs, or feature-based methods with $\pi_\mathcal{S}, \pi_\mathcal{T}$ defined as the intermediate activation maps.
 Response-based KD methods generally keep the same peak memory with the student model, since they don't alter the student architecture.
 For feature-based KD methods, some might increase the student's peak memory due to extra trainable modules.
 However, none of these KD methods can lead to peak memory lower than the student, which is a lower bound.
 Our method reaches such a lower bound and achieves the highest accuracy compared to existing KD methods, as shown in our comprehensive experiments.}

 \textbf{Denoising Diffusion Probabilistic Models}\quad Diffusion models~(\cite{ddpm2020}) are latent variable models of the form $p_\theta(x_0):=\int p_\theta(x_{0:T})d{x_{1:T}}$, where $x_1, ... , x_T$ are latents of the same dimensionality as the data $x_0\sim q(x_0)$.
 The joint distribution $p_\theta(x_{0:T})$ is defined as a Markov chain with learned Gaussian transition starting at $p(x_T)=\mathcal{N}(x_T;\textbf{0},\textbf{I})$, where $T$ is the maximum time step.
 In the training process, we are given a noisy input $x_t$, which is derived from the data $x_0$ and noise $\epsilon\sim\mathcal{N}(\textbf{0},\textbf{I})$:
 \begin{small}
 \begin{eqnarray}\label{eq:ddpm input}
 x_t = \sqrt{\bar{\alpha_t}}x_0 + \sqrt{1-\bar{\alpha_t}}\epsilon,
 \end{eqnarray}
 \end{small}where $\bar\alpha_t := \prod_{s=1}^t(1-\beta_s)$, and $\beta_s$ is the forward process variances fixed as constant in DDPM~(\cite{ddpm2020}).
 The loss of diffusion model is generally defined as follows:
 \begin{small}
 \begin{eqnarray}\label{eq:ddpm loss}
 \mathcal{L}_{diff} = ||\epsilon-\epsilon_\theta(x_t,t)||_2^2.
 \end{eqnarray}
 \end{small}
 
 The noisy input $x_t$, accompanied by a time step embedding $t$, is input into a denoising autoencoder, specifically a U-Net network as in~\cite{ddpm2020}, to estimate the noise component $\epsilon_\theta(x_t, t)$.

\subsection{Proposed Distillation Framework}
\label{sec: method}

\begin{figure*}[t!]
\centerline{\includegraphics[scale=0.7]{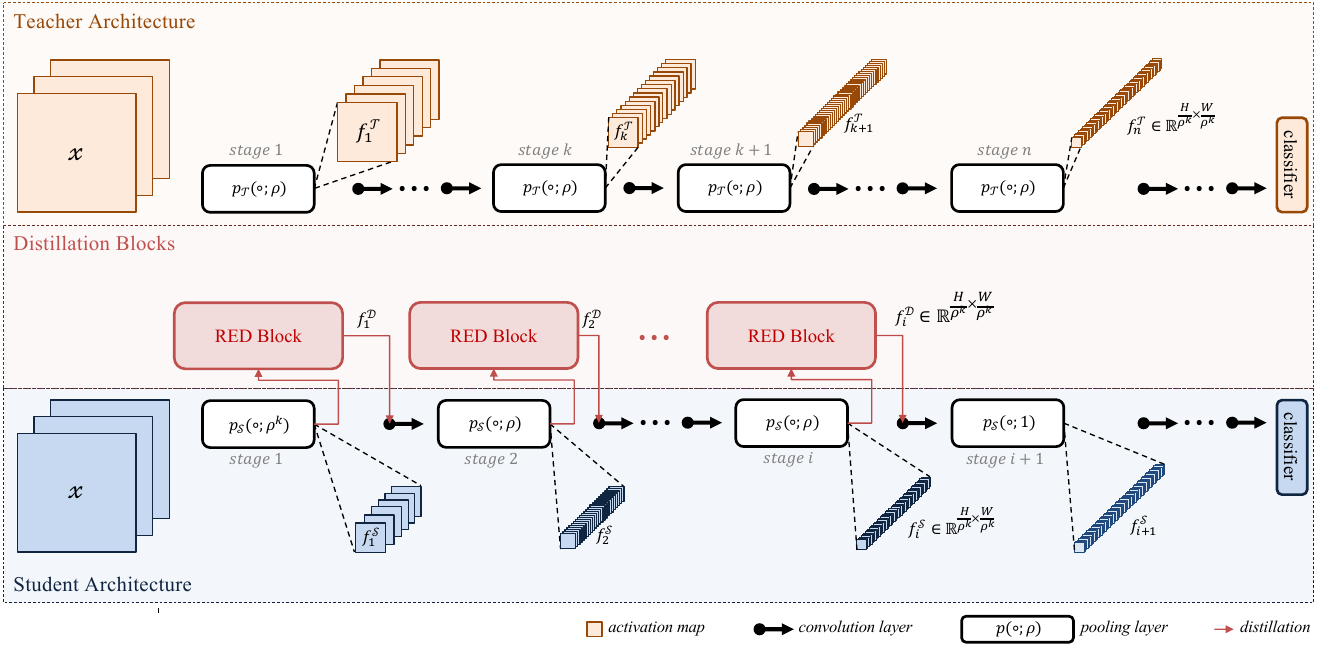}}
 \caption{Our proposed residual encoded distillation framework (ReDistill).
 RED blocks are incorporated into the student model following the pooling layers to minimize the discrepancy between the down-sampled features of the student and teacher models.
 }
\label{fig: proposed method overview}
\end{figure*}

Our proposed framework is illustrated in Fig.~\ref{fig: proposed method overview}.
 To reduce the activation peak memory, the initial pooling layer of the student is assigned a larger pooling stride.
 However, we still keep the same spatial dimensions of the activations to the final fully connected layer for the teacher and student.
 Thus, the student has fewer pooling layers but with a larger pooling stride at the initial pooling layer compared to the teacher.

\begin{figure}
\centerline{\includegraphics[scale=0.5]{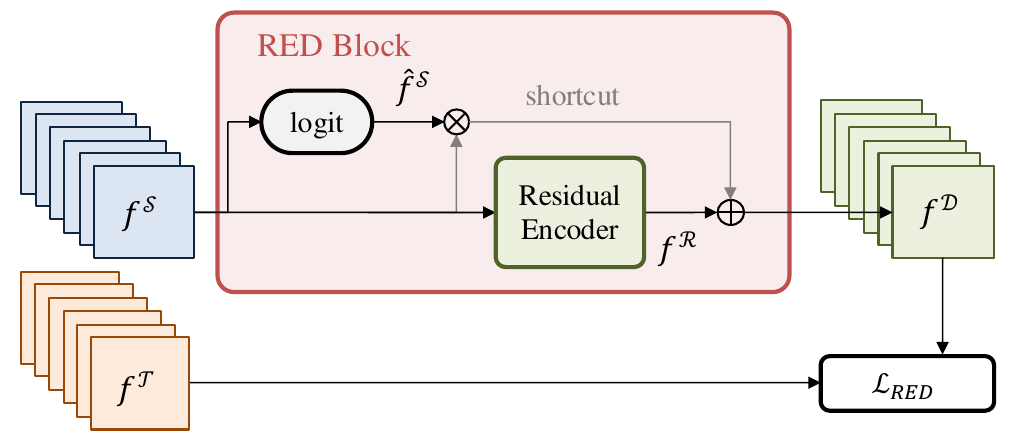}}
\caption{Residual Encoded Distillation (RED) Block. We use a logit module for the multiplicative gating mechanism and a residual encoder for additive residual learning.
 }
\label{fig: residual encoded distillation}
\end{figure}

 Take an input image $x\in\mathbb{R}^{H\times W\times C}$ as the example.
 The teacher and the student network are divided into several stages by pooling layers as shown in Fig.~\ref{fig: proposed method overview}.
 We assume that all pooling layers of the teacher have the same pooling stride $\rho$ for simplicity.
 The initial pooling layer of the student at stage $1$ is assigned with a stride $\rho^k$. Hence, the student feature map $f^\mathcal{S}_1\in\mathbb{R}^{\frac{H}{\rho^k}\times{\frac{W}{\rho^k}\times C_1}}$ at the output of this pooling layer has the same spatial dimension as the teacher feature map $f^\mathcal{T}_{k}\in\mathbb{R}^{\frac{H}{\rho^k}\times{\frac{W}{\rho^k}\times C_k}}$ at the output of the $k$-th pooling layer.
 Then the matched feature maps $f^\mathcal{S}_1$ and $f^\mathcal{T}_k$ will be fed into the residual encoded distillation (RED) block illustrated in Fig.~\ref{fig: residual encoded distillation}, to compute the distillation loss.
 The output $f^D_1$ is to be fed into the following layers of the student network.

 Similarly, the feature map $f^\mathcal{S}_2$ of the student at stage $2$ also has the same spatial dimension as the feature map $f^\mathcal{T}_{k+1}$ of the teacher at stage $k+1$.
 They are fed into another RED block to calculate distillation loss and output $f^D_2$ to the following layers.
 This process is repeated until the last pooling layer $p_\mathcal{T}(\circ;\rho)$ at stage $n$ of the teacher.
 We assume the size of the output feature map $f^\mathcal{T}_{n}$ of the teacher is identical to the output feature map $f^\mathcal{S}_i$ of the student at stage $i$, where $i-1+k=n$.
 The following pooling layers of the student at stages $i+1, i+2, ..., n$ are all assigned with stride $1$.
 Hence, these pooling layers do not change the spatial dimension and are similar to standard convolution.
 As a result, the final aggregated features of the student and the teacher have the same spatial dimensions.
 \rebuttal{In Fig.~\ref{fig: network config}, we show an example of ResNet18 with aggressive pooling and how to integrate our proposed RED blocks.}


\subsection{Residual Encoded Distillation Block}
\label{sec:red}

 The proposed Residual Encoded Distillation (RED) Block, depicted in Fig.~\ref{fig: residual encoded distillation}, serves as the central module of our framework. It is designed to ensure that the output of a pooling layer closely resembles the distribution of the input while preserving essential features at a reduced spatial dimension.
 To accomplish this, the RED Block is designed to be lightweight, enabling it to modify the feature space distribution of the student's pooling layer effectively. This allows the student model to learn the down-sampled features of the teacher model.
 Meanwhile, this block introduces non-linearity to the pooling layer, enabling the student's pooling layer to aggregate features like a standard pooling layer and adjust the feature distribution similar to a convolutional layer with an activation function.
 Specifically, we use a logit module for the multiplicative gating mechanism and a residual encoder for additive residual learning.
 The residual encoded distillation block could be formulated as follows:
 \begin{small}
 \begin{eqnarray}
  f^\mathcal{D} &=&  f^\mathcal{R} + f^\mathcal{S}*\hat{f}^\mathcal{S}, \label{eq:residual encoded distillation block} \\
 ResidualEncoder &=& ReLU6(BN(Conv_{3\times3}(\cdot))), \\
 logit &=& Sigmoid(BN(Conv_{1\times1}(\cdot))),
 \end{eqnarray}
 \end{small}where $f^\mathcal{S}$ is the feature map from the student model.
 The logit module consists of a $1\times1$ convolution layer, a batch norm layer, and a sigmoid activation function, generating element-wise weights $\hat{f}^\mathcal{S}$ like a gate to suppress non-significant components of $f^\mathcal{S}$.
 The residual encoder module consists of a $3{\times}3$ convolution layer, a batch norm layer, and a relu-6 activation function, yielding the residual item $f^\mathcal{R}$.
 \rebuttal{
 We use relu-6 to bound activations to prevent exploding gradients.
 Besides, relu-6 is widely used in efficient neural network design~(\cite{mobilenetv2, mobilenetv3, mcunet}) since it is particularly useful for fixed-point or low-precision inference in quantization.
 For kernel size in RED module, we choose $1 \times 1$ kernel for the logit module (LM), because LM is a gating mechanism to select critical components of the activations.
 We choose BatchNorm layer because the proposed framework is primarily used for the distillation of CNNs, for which BatchNorm is commonly used.}
 We hypothesize that the output of the student pooling layers might lack some crucial information compared to the teacher's down-sampled features, which could be compensated by the residual item $f^\mathcal{R}$.

\subsection{Loss Function}
\label{sec: loss function}
 The RED loss first calculates the mean value alongside the channel dimension and then minimizes the cosine distance between the teacher's feature map $f^\mathcal{T}$ and the RED block's output $f^\mathcal{D}$, which is formulated as:
 \begin{small}
 \begin{eqnarray}\label{eq:RED block loss}
 \mathcal{L}_{RED}(f^\mathcal{T}, f^\mathcal{D}) = d\{\frac{\sum_{c\in C_\mathcal{T}}{f^\mathcal{T}_c}}{|C_\mathcal{T}|}, \frac{\sum_{c\in C_\mathcal{S}}{f^\mathcal{D}_c}}{|C_\mathcal{S}|}\},
 \end{eqnarray}
 \end{small}where $f^\mathcal{D}$ is calculated by Equation~\ref{eq:residual encoded distillation block}, and $d$ denotes the distance measurement.
 $C_\mathcal{T}$ and $C_\mathcal{S}$ denote the channel dimension of teacher's feature map $f^\mathcal{T}$ and the RED block's output $f^\mathcal{D}$, respectively.
 The final loss function of the proposed method is as follows:
 \begin{small}
 \begin{eqnarray}\label{eq: final loss}
 \mathcal{L} = \mathcal{L}_{task} + \sum_{i=1}^I{\alpha\mathcal{L}_{RED_i}},
 \end{eqnarray}
 \end{small}where $\mathcal{L}_{task}$ is the vanilla loss from the task, such as the Binary-Cross-Entropy (BCE) loss for image classification. $I$ denotes the number of RED blocks.
 $\alpha$ is an experimental hyper-parameter to scale RED loss.

\subsection{Distillation for Diffusion Model}
\label{sec: distillation for diffusion model}


\begin{figure}
\centerline{\includegraphics[scale=0.7]{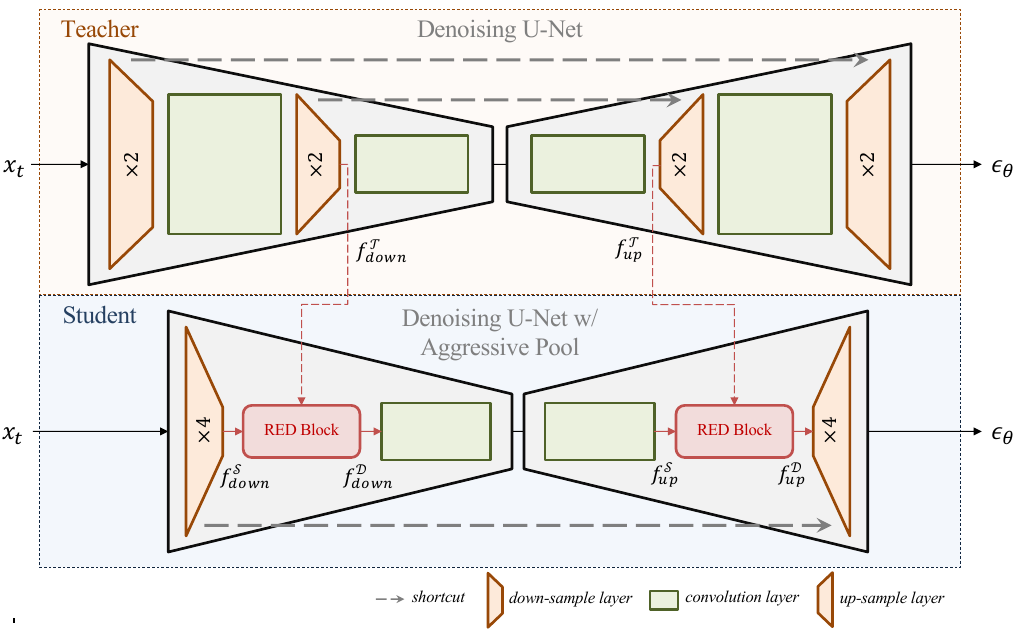}}
\caption{ReDistill for denosing network in DDPM~(\cite{ddpm2020}).
 We integrate RED blocks into the student model after the down-sampling layer in the encoder and before the up-sample layer in the decoder.
 }
\label{fig: distillation for diffusion}
\vspace{-3mm}
\end{figure}

 In this section, we introduce how to integrate the proposed distillation framework into a U-Net based denoising network described in DDPM~(\cite{ddpm2020}), as shown in Fig.~\ref{fig: distillation for diffusion}.
 For convenience, let's assume the teacher model is a U-Net with two down-sample layers, each having a stride of $2$, while the student is a U-Net with the aggressive pooling setting, i.e., it just has one down-sample layer with the stride $4$.
 We use the output $f_{down}^\mathcal{S}$ from the down-sample layer of the student model and the output $f_{down}^\mathcal{T}$ from the second down-sample layer of the teacher model, as they share the same spatial dimension.
 $f_{down}^\mathcal{S}$ and $f_{down}^\mathcal{T}$ are input into a RED block incorporated into the student model, producing the output $f_{down}^\mathcal{D}$, which is then fed into the subsequent convolutional layers.
 Symmetrically, the student model has one up-sample layer with an expansion ratio of $\times4$, while the teacher model has two up-sample layers, each with an expansion ratio of $\times2$.
 The input $f_{up}^\mathcal{S}$ of the student's up-sample layer and the input $f_{up}^\mathcal{T}$ of the teacher's first up-sample are input into another RED block, producing the output $f_{up}^\mathcal{D}$.
 This output replaces the $f_{up}^\mathcal{S}$ as the new input of the student's up-sample layer.
 For DDPM distillation, the loss function is defined as Equation~\ref{eq: final loss}, while the $\mathcal{L}_{task}$ is replaced by $\mathcal{L}_{diff}$ defined in Equation~\ref{eq:ddpm loss}.

\section{Experiments}
\label{experiments}
 In Sections~\ref{sec: datasets} and~\ref{sec: implementation details}, datasets and implementation details for image classification and image generation tasks are introduced separately.
 In Sections~\ref{sec: image classification} and~\ref{sec: image generation}, we conduct experiments on various vision tasks to illustrate the effectiveness of the proposed method with state-of-the-art distillation methods under different backbone architectures and datasets.
 We also compare the memory footprint of our method with the teacher and the student architectures, and deploy our method on the edge device.
 Results are reported in Section~\ref{sec: analyses}.
 At last, some ablation study related with module discussion, loss function, and distillation strategy are reported in Section~\ref{sec: ablation study}.

\subsection{Datasets}
\label{sec: datasets}

 \paragraph{Datasets for Image Classification} 1) STL10~(\cite{coates2011analysis}) contains 5K training images with 10 classes and 8K testing images of resolution $96\times96$ pixels. Specifically, we resize the image resolution to $128 \times 128$ pixels for aggressive pooling. 2) ImageNet~(\cite{ILSVRC15}) is a widely-used dataset of classification, which provides 1.2 million images for training and 50K images for validation over 1,000 classes. We keep the same resolution of $224 \times 224$ pixels as the origin for aggressive pooling.

 \paragraph{Datasets for Image Generation} 1) CIFAR-10~(\cite{alex2009learning}) comprises 60,000 color images of 32x32 resolution across 10 classes, with each class containing 6,000 images. The dataset is divided into 50,000 training images and 10,000 test images.
 We keep the original resolution of $32\times32$ in our experiments.
 2) Celeb-A~(\cite{liu2015faceattributes}) is a large-scale face attributes dataset containing over 200,000 celebrity images, each annotated with 40 attributes.
 We use the resized resolution of $64\times64$, which is widely used in diffusion-based methods~(\cite{ddim2020, uvit2023}) in our experiments.

 \subsection{Implementation Details and Baselines}
 \label{sec: implementation details}

 \paragraph{Details for Image Classification} Unlike traditional distillation tasks, we only modify the pooling layer strides instead of the depth and width of the network to get the student model, which is called the aggressive pooling setting.
 The advantage of the aggressive pooling setting is to reduce the peak memory and also reduce the computational complexity and inference time of the network. 
 Specifically, we increase the first pooling layer stride $\times 2 \sim \times 8$ times and adjust the last several pooling layer strides to ensure the final output of the student model has the same information density as the teacher model.
 All experiments are implemented in Pytorch and evaluated on 4 NVIDIA A100 GPUs.

 On STL10 dataset, we experiment with three representative and widely-used network architectures, including MobileNetV2~(\cite{mobilenetv2}), MobileNetV3~(\cite{mobilenetv3}), and ResNext~(\cite{resnext}).
 The proposed method is compared with several distillation methods~(\cite{KD2015, FITNET2015, AT2016, Similarity2019, VID2019, RKD2019, Abound2018, Factor2020, NST2017}).
 Specifically, the student architecture is trained from scratch as being distilled from pre-trained teacher architecture by different methods for 300 epochs.
 The batch size is set to 8 and the dropout rate is set to $0.2$.
 The SGD with momentum equal to 0.9 is used as the optimizer.
 The initial learning rate is set to 0.01, which is reduced by a factor $0.2$ at the $180^{th}$, $240^{th}$ and $270^{th}$ epoch, respectively.
 The $\alpha$ in Equation ~\ref{eq: final loss} is set to $50$.

 On ImageNet dataset, we experiment on the MobileNetV2~(\cite{mobilenetv2}), and ResNet~(\cite{he2016deep}), which are widely used in distillation benchmarks~(\cite{MLLD2023, ReviewKD2021, crd2020}).
 The proposed method is compared with response-based methods like KD~(\cite{KD2015}), DKD~(\cite{DKD2022}), MLLD~(\cite{MLLD2023}), \rebuttal{LSKD~(\cite{lskd})}, and feature-based methods like FitNet~(\cite{FITNET2015}), RKD~(\cite{RKD2019}), ReviewKD~(\cite{ReviewKD2021}), CRD~(\cite{crd2020}), and \rebuttal{OFAKD~(\cite{ofakd})}.
 All these methods are widely used in knowledge distillation and, to the best of our knowledge, yield SOTA performance.
 We use the same experiment settings as~(\cite{MLLD2023}) but keep training for 300 epochs and decay the learning rate at the $180^{th}$, $240^{th}$ and $270^{th}$ epoch with factor $0.1$, since the student architectures with the aggressive pooling setting generally require more epochs to converge.
 The $\alpha$ in Equation ~\ref{eq: final loss} is set to be $1$.

 \paragraph{Details for Image Generation} For the teacher model, we keep the same experiment settings as DDPM~(\cite{ddpm2020}) with applying $T=1000$, $\beta_1=10^{-4}$, $\beta_T=0.02$, and the U-Net backbone with $4$ different feature map resolutions ($32\times32$ to $4\times4$ for CIFAR-10, while $64\times64$ to $8\times8$ for Celeb-A).
 For the student model, we increase the first pooling layer stride of the U-Net backbone $\times2$ times while adjusting the last pooling layer stride to keep the same latent feature resolution.
 The same stride modification is symmetrically applied to the up-sample layers of the U-Net, and thus with $3$ different feature map resolutions ($16\times16$ to $4\times4$ for CIFAR-10, while $32\times32$ to $8\times8$ for Celeb-A).
 For our method, the RED blocks are inserted after not only the down-sample layers but also the up-sample layers. 
 For CIFAR-10 dataset, we train all models for 1000K iterations and sample 50K images for FID~(\cite{heusel2018gans}) \& IS~(\cite{salimans2016improved}) evaluation.
 For Celeb-A dataset, we train all models with 250K iterations and sample 50K images for FID~(\cite{heusel2018gans}) \& IS~(\cite{salimans2016improved}) evaluation.
 All experiments are implemented in Pytorch and evaluated on an NVIDIA 4090 GPU.

\vspace{-6mm}
\rebuttal{\paragraph{Details for Theoretical Peak Memory} For the theoretical peak memory analysis, we follow previous work, MCUNet~(\cite{mcunet, mcunetv2, mcunetv3}), for calculating peak memory.
 Generally, it traces the memory consumption for each layer.
 For the standard convolutional layer, the maxpool layer, or the average pool layer, the memory consumption is the summation of the input activation memory size and output activation memory size.
 For a group convolutional layer, the memory consumption is the summation of the input activation memory size, the output activation memory size, and a buffer with size equal to one channel convolutional kernel in the group.
 For the residual connection in the network, the residual activations memory size will be added into memory tracing until the residual item is added with the output activations.
 \vspace{-6mm}
 }

\begin{figure}
\centerline{\includegraphics[scale=1.0]{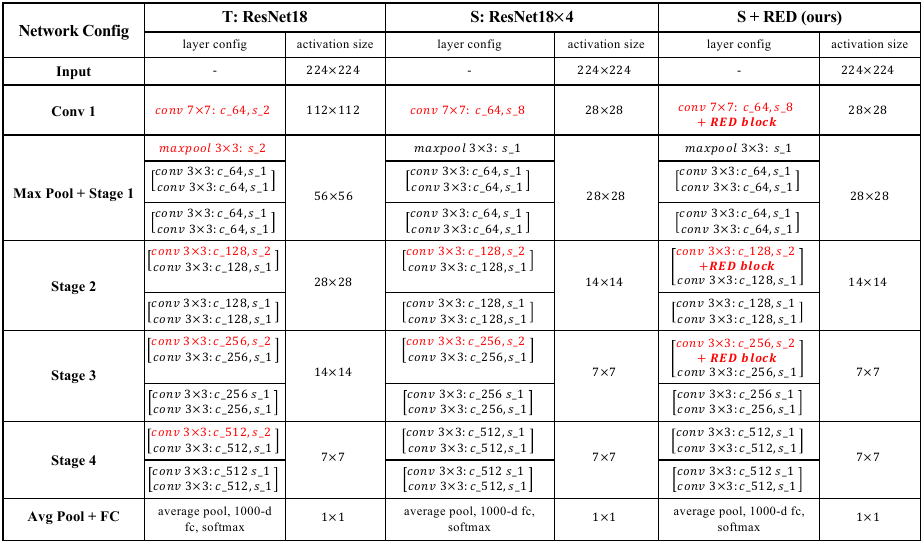}}
\caption{
\rebuttal{Example of our proposed Aggressive Pooling Setting with ResNet18.
 We highlight all downsampling layers (either conv layer or maxpool layer with the stride larger than 1) in red color.
 For example, in the `Conv 1' cell of `T: ResNet18', `conv $7 \times 7: c\_64, s\_2$' denotes this convolution layer is with kernel size $7 \times 7$, number of channels 64, and stride 2.
 In the following row, `maxpool $3 \times 3: s\_2$' denotes the max pool layer with kernel size $3 \times 3$ and stride 2.
 In the student network `S: ResNet18$\times 4$', we increase the stride of the first downsampling layer `Conv 1' to $4\times$ from $s\_2$ to $s\_8$, while setting the stride to 1 of the maxpool layer and the last downsampling layer (i.e., the first conv layer in Stage 4), in order to get the same activation size before average pool and fc layer.}
}
\label{fig: network config}
\vspace{-4mm}
\end{figure}

\rebuttal{\paragraph{Details for Aggressive Pooling Setting.} To explain how exactly the image classification models are configured, we list the network config details for the `T: ResNet18' and `S: ResNet18$\times 4$' distillation pair that we used in the third column of Table~\ref{tab: comparison on imagenet}.
 As shown in Fig.~\ref{fig: network config}, we aggressively increase the stride of the first downsampling layer `Conv 1' from 2 to 8 following the aggressive pooling setting, while setting the stride to 1 of maxpool layer and the last downsampling layer, i.e., the first conv layer in `Stage 4', for keeping the same activation size for average pool and fc layer.
 Then the student network possesses only three downsampling layers less than five downsapling layers of the teacher network.
 We integrate the proposed RED block after each downsampling layer of the student in our ReDistill framework to improve the student's performance while maintaining low peak memory.}

 \subsection{CNN-based Image Classification}
 \label{sec: image classification}

 \begin{table*}[tphb!]
\caption{Top-1 accuracy (\%) on ImageNet. The $3^{rd}$ and $4^{th}$ columns show the results with identical teacher and student architecture families, while the $5^{th}$ column shows the results with different teacher and student architectures.}
\label{tab: comparison on imagenet}
\begin{center}
\setlength{\tabcolsep}{2mm}{
\begin{threeparttable}
\small
\begin{tabular}{l|c|c|c|c}
\hline
\hline

\multirow{4}{*}{Method} & Teacher & ResNet18 & ResNet50 & ResNet152 \\
& Top1 Acc. (\%) & 69.75 & 76.13 & 78.32 \\
\cline{2-5}
& Student & ResNet18$\times 4$ & ResNet50$\times 4$ & MbNetV2$\times 2$ \\
& Top1 Acc. (\%) & 61.79 & 69.50 & 62.65 \\

\hline
 
\multirow{4}{*}{Response} & KD~(\cite{KD2015}) & 63.63 &  70.60 & 62.85 \\ 
 & DKD~(\cite{DKD2022}) & 63.22 & - & 66.27 \\ 
 & MLLD~(\cite{MLLD2023}) & 64.66 & 70.77 & 68.36 \\ 
 & \rebuttal{LSKD~(\cite{lskd})} & \rebuttal{63.81} & \rebuttal{72.28} & \rebuttal{66.21} \\

\hline

\multirow{6}{*}{Feature} & FitNet~(\cite{FITNET2015}) & 62.13 & 71.77 & 60.79 \\ 
 & RKD~(\cite{RKD2019}) & 61.49 & 66.88 & - \\
 & ReviewKD~(\cite{ReviewKD2021}) & 63.30 & 70.22 & 63.07 \\ 
 & CRD~(\cite{crd2020}) & 64.01 & 71.07 & 65.60 \\ 
 & \rebuttal{OFAKD~(\cite{ofakd})} & \rebuttal{64.83} & \rebuttal{-} & \rebuttal{68.49} \\

\cline{2-5}

& \textbf{RED (ours)} & \textbf{65.23} & \textbf{73.23} & \textbf{68.89} \\ 

\hline
\hline
\end{tabular}
    \footnotesize{
    `-' denotes that we do not get reasonable results for the student architecture with the distillation method under the aggressive pooling setting.
    }
\end{threeparttable}
}
\end{center}
\end{table*}

Table~\ref{tab: comparison on imagenet} shows the results on the ImageNet~ (\cite{ILSVRC15}) dataset, with the setting that the teacher model and student model are in identical architecture families or different architectures.
 `$\times n$' denotes that we increase the $1^{st}$ pooling layer stride of this architecture with $n$ times, and the best results are highlighted in boldface.
 Our method achieves the best performance compared with different response-based and feature-based distillation methods, no matter for identical architecture family knowledge distillation, as shown in the $3^{rd}$ and $4^{th}$ columns of Table~\ref{tab: comparison on imagenet}, or different architecture knowledge distillation, as shown in the $5^{th}$ column of Table~\ref{tab: comparison on imagenet}.

\begin{table*}[b!]
\caption{{Top-1 accuracy (\%) on STL10 with identical teacher and student architecture family.} 
}
\label{tab: stl10 homogeneous}
\begin{center}
\setlength{\tabcolsep}{2mm}{
\begin{threeparttable}
\small
\begin{tabular}{l|c|c|c|c}
\hline
\hline

 \multirow{4}{*}{Method} & T: MbNetV2 & T: MbNetV3-Small & T: ResNext18 & \multirow{4}{*}{\makecell[c]{Distill Time\\(s/epoch)}}\\ 
 & 85.34 & 83.74 & 85.12 \\ 
\cline{2-4}
 & S: MbNetV2$\times 4$ & S: MbNetV3-S$\times 4$ & S: ResNext18$\times 4$ \\ 
 & 76.18 & 71.27 & 79.07 \\ 

\hline

 KD~(\cite{KD2015}) & 79.26 & 71.56 & 81.27 & \textbf{3.23} \\ 

 FitNet~(\cite{FITNET2015}) & 78.55 & 71.93 & 80.21 & 3.28 \\ 
 AT~(\cite{AT2016}) & 81.35 & 75.88 & 82.90 & 3.33 \\ 
 SP~(\cite{Similarity2019}) & - & 72.42 & 77.59 & 3.31 \\ 
 VID~(\cite{VID2019}) & 77.53 & 71.83 & 76.99 & 3.40 \\ 
 RKD~(\cite{RKD2019}) & 78.00 & 67.27 & 75.80 & 3.33 \\ 
 AB~(\cite{Abound2018}) & 80.79 & 72.94 & 81.42 & 3.69 \\ 
 FT~(\cite{Factor2020}) & 74.74 & 72.11 & 80.64 & 3.35 \\ 
 NST~(\cite{NST2017}) & 82.66 & 76.11 & 81.89 & 9.33 \\ 

\hline

 \textbf{RED (ours)} & \textbf{83.97} & \textbf{77.31} & \textbf{84.80} & 3.88 \\ 

\hline
\hline
\end{tabular}
    \footnotesize{
    `-' denotes that we do not get reasonable results for the student architecture with the distillation method under the aggressive pooling setting.
    }
\end{threeparttable}
}
\end{center}
\end{table*}

\begin{table*}[h]
\caption{{Top-1 accuracy (\%) on STL10 with different teacher and student architecture.} 
}
\label{tab: stl10 heterogeneous}
\begin{center}
\setlength{\tabcolsep}{2mm}{
\begin{threeparttable}
\small
\begin{tabular}{l|c|c|c}
\hline
\hline

 \multirow{4}{*}{Method} & T: ResNext18 & T: ResNext18 & T: MbNetV2 \\ 
 & 85.12 & 85.12 & 85.34 \\ 
\cline{2-4}
 & S: MbNetV2$\times 4$ & S: MbNetV3-Small$\times 4$ & S: MbNetV3-Small$\times 4$ \\ 
 & 76.18 & 71.27 & 71.27 \\ 

\hline

 KD~(\cite{KD2015}) & 78.14 & 69.98 & 70.95 \\ 

 FitNet~(\cite{FITNET2015}) & 80.46 & 73.23 & 72.88 \\ 
 AT~(\cite{AT2016}) & 80.63 & 74.58 & 73.39 \\ 
 SP~(\cite{Similarity2019}) & 67.56 & 63.91 & - \\ 
 VID~(\cite{VID2019}) & 74.40 & 69.19 & 71.69 \\ 
 RKD~(\cite{RKD2019}) & 70.63 & 72.10 & 68.78 \\ 
 AB~(\cite{Abound2018}) & 81.46 & 73.98 & 75.20 \\ 
 FT~(\cite{Factor2020}) & 77.33 & 69.79 & 66.85 \\ 
 NST~(\cite{NST2017}) & 79.09 & 65.04 & 73.26 \\ 

\hline

 \textbf{RED (ours)} & \textbf{83.23} & \textbf{77.15} & \textbf{77.19} \\ 

\hline
\hline
\end{tabular}
    \footnotesize{
    `-' denotes that we don't get reasonable results for the student architecture with the distillation method under the aggressive pooling setting.
    }
\end{threeparttable}
}
\end{center}
\vspace{-3mm}
\end{table*}


 \rebuttal{
 We find some distillation methods perform even worse than the student model itself without any distillation since these methods are not specially designed for the aggressive pooling setting and would be sensitive to the resolution of the feature maps, like FitNet, or require multi-scale feature maps, like RKD.
 Generally, conventional feature-based methods match the activations of the teacher model and student model stage by stage.
 In each stage, these activations possess the same resolution.
 However, in our aggressive pooling setting, the resolution of student activations and teacher activations are mismatched in each stage, and thus, traditional distillation methods are not guaranteed to be positively effective in knowledge distillation with aggressive pooling.
 This observation also illustrates the necessity of the proposed ReDistill framework.
 }

 Table~\ref{tab: stl10 homogeneous} shows the classification results on STL10 dataset with the setting that the teacher model and student model are in identical architecture family.
 Same as ImageNet, `$\times n$' denotes we increase the $1^{st}$ pooling layer stride of this architecture with $n$ times, and the best results are highlighted in boldface.
 \rebuttal{Some KD methods in Table~\ref{tab: comparison on imagenet} require a customized dataloader with a contrastive version of the input data, like CRD~\cite{crd2020}, MLLD~\cite{MLLD2023}.
 However, they don’t provide the corresponding dataloader for STL10 dataset in their original code implementations.
 Thus, we compare our method on STL10 dataset with some feasible KD methods in Table~\ref{tab: comparison on imagenet}, like KD, FitNet and RKD, and also compare with some other commonly-used KD methods, like AT, VID, NST, etc.
 }
 Under the identical architecture family setting, our method performs the best among all state-of-the-art distillation methods.
 We also measure the average distillation time for different distillation methods, as shown in the last column of Table~\ref{tab: stl10 homogeneous}.
 Our method achieves similar time consumption compared to most distillation methods.
 For different architecture knowledge distillation, as shown in Table~\ref{tab: stl10 heterogeneous}, our method achieves better performance compared with other distillation methods as well.

\subsection{Image Generation}
\label{sec: image generation}

\begin{wraptable}{r}{0.52\textwidth}
\vspace{-5mm}
\caption{Results on U-Net~(\cite{ronneberger2015u}) based DDPM~(\cite{ddpm2020}).}
\vspace{-2mm}
\label{tab: ddpm results}
\begin{center}
\setlength{\tabcolsep}{1.2mm}{
\begin{threeparttable}
\small
\begin{tabular}{l|l|c|c}
\hline
\hline
Dataset & Method & IS$\uparrow$ & FID$\downarrow$ \\
\hline
\multirow{4}{*}{CIFAR10} & T: U-Net w/ DDPM & \rebuttal{\textbf{9.70 $\pm $0.13}} & \rebuttal{\textbf{3.75}} \\
& S: U-Net$\times2$ w/ DDPM & \rebuttal{9.12 $\pm$ 0.14} & \rebuttal{14.28} \\
& \rebuttal{S + MSE-Distill} & \rebuttal{9.62 $\pm$ 0.11} & \rebuttal{12.04} \\
& \textbf{S w/ RED (ours)} & \rebuttal{\textbf{9.63 $\pm$ 0.10}} & \rebuttal{\textbf{10.85}} \\ 
\hline
\multirow{4}{*}{Celeb-A} & T: U-Net w/ DDPM & \textbf{2.97 $\pm$ 0.04} & \textbf{19.61}  \\
& S: U-Net$\times2$ w/ DDPM & 2.77 $\pm$ 0.02 & 23.03 \\
& \rebuttal{S + MSE-Distill} & 2.71 \rebuttal{$\pm$ 0.02} & \rebuttal{22.34} \\
& \textbf{S w/ RED (ours)} & \textbf{2.88 $\pm$ 0.03} & \textbf{21.16} \\ 

\hline
\hline
\end{tabular}
\end{threeparttable}
}
\end{center}
\vspace{-3mm}
\end{wraptable}

\paragraph{Quantitative Results} Table~\ref{tab: ddpm results} shows the results on CIFAR10~(\cite{alex2009learning}) dataset and Celeb-A~(\cite{liu2015faceattributes}) dataset.
 Our method reduces the fidelity degradation of the student model with a first pooling stride that is twice that of the teacher model.
 \rebuttal{Specifically, our method achieves $3.43$ lower FID and $0.51$ higher IS score than the student model on CIFAR10 dataset, while $1.6$ lower FID and $0.11$ higher IS score than the student model on Celeb-A, respectively.
 We also compare to a simple distillation method, which we termed MSE-Distill-DDPM, by directly matching the final output activations of the teacher model and student model using MSE loss.
 On CIFAR10 dataset, the student with MSE-Distill achieves 12.04 FID and 9.62 IS score, while our method achieves a better 10.85 FID and 9.63 IS score.}
 Due to implementation specifics, the models are not augmented with EMA~(\cite{hunter1986exponentially}), leading to results slightly different from those in the original paper.

\begin{figure}[h]
\centerline{\includegraphics[scale=0.58, trim={0 0 244 0},clip]{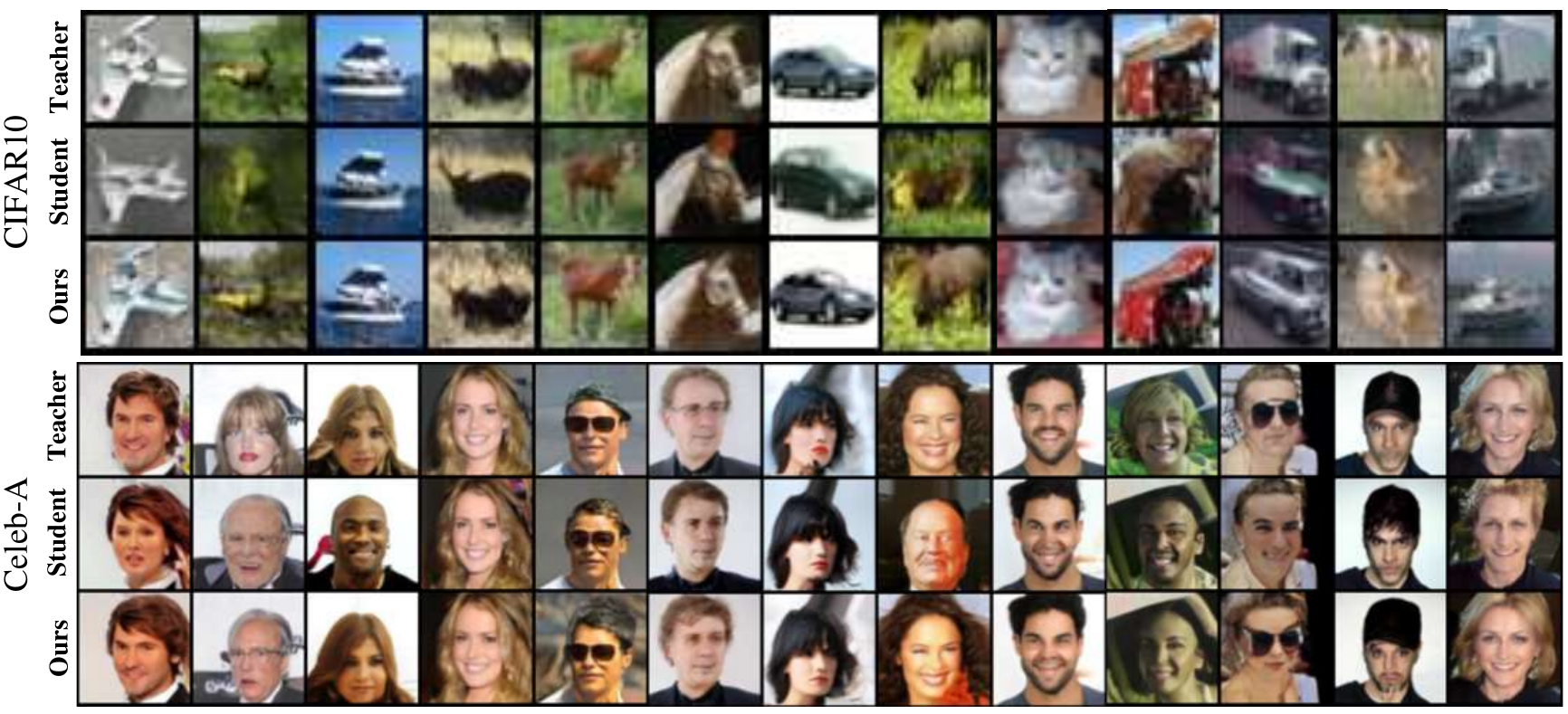}}
\caption{Generated images on CIFAR10 and Celeb-A with the same noise item $\epsilon_t$ for all models. 
 Generally, our results are semantically closer to the teacher's.
 }
\label{fig: visualization}
\end{figure}

\begin{figure}[htbp]
\centerline{\includegraphics[scale=0.398]{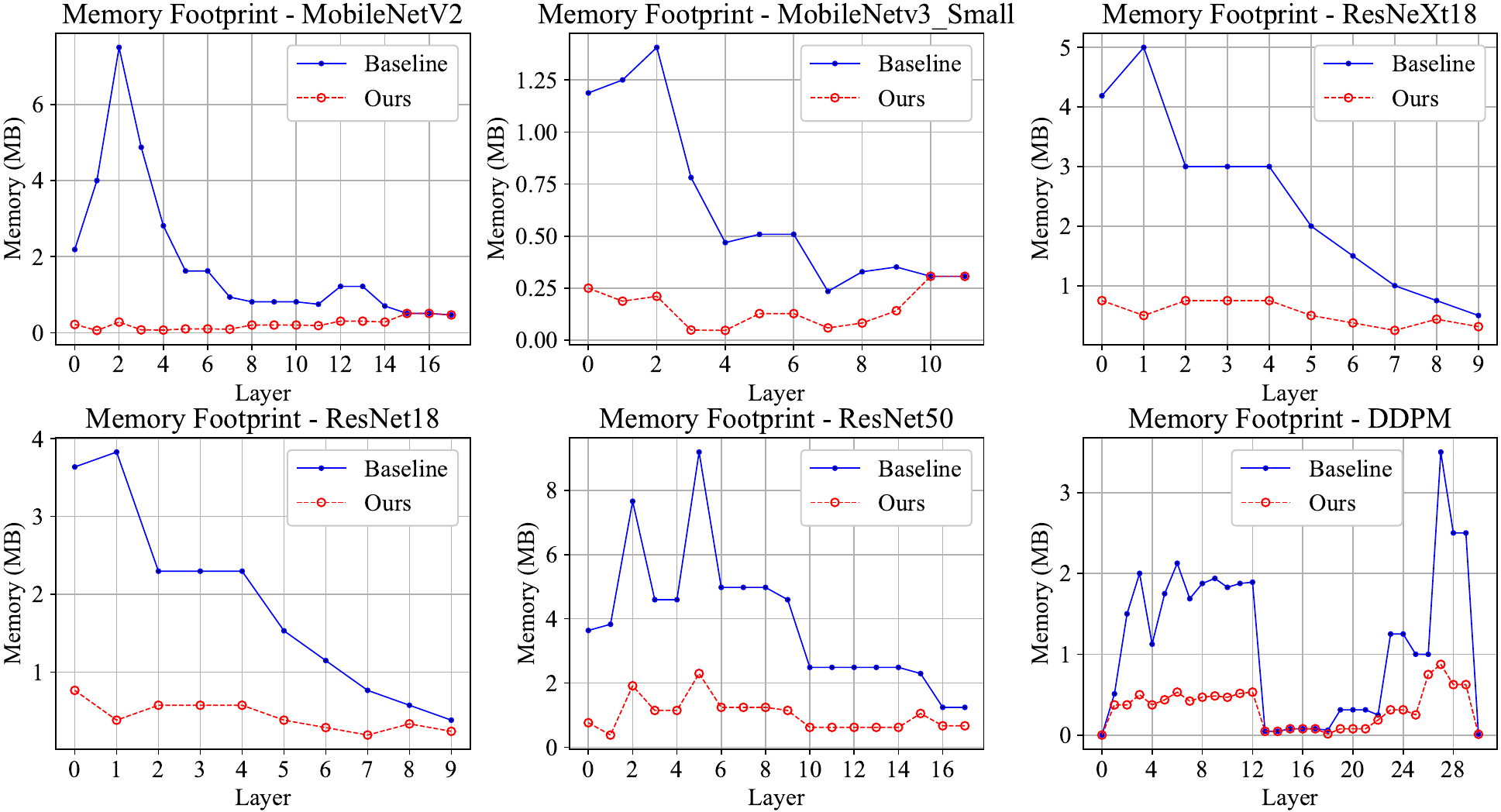}}
\caption{Memory footprint for each layer in teacher and student networks (top-all: STL10 classification task; bottom-left, bottom-mid: ImageNet classification task; bottom-right: Celeb-A generation task).
The student model is the teacher model's modified version of being assigned with the aggressive pooling strategy and enhanced with our proposed RED blocks, representing our method.
 }
\label{fig: memory footprint}
\end{figure}

 \paragraph{Visualization} To further illustrate how the proposed method improves the fidelity of image generation, we visualize some samples generated by the teacher model, the student model, and our method.
 Specifically, we utilize the same noise item $\epsilon_t$ in each time step for all these three models.
 In this way, the generated images are expected to be visually similar if the two models have comparable capabilities.
 As shown in Fig.~\ref{fig: visualization}, generally, the images generated by our method are semantically closer to those generated by the teacher model, although in some cases, they are closer to those generated by the student model.
 Due to the high capability of the teacher model, our method, as an intermediate model from the teacher to the student, achieves higher fidelity than the student model.

\subsection{Memory Footprint and Edge Device Deployment}
\label{sec: analyses}



 \paragraph{Memory Footprint} To intuitively demonstrate the peak memory reduction enabled by our proposed method, we trace the memory footprint in layer-wise for all identical architecture family teacher-student pairs, as shown in Fig.~\ref{fig: memory footprint}.
 Compared to CNN models with vanilla pooling, our method with aggressive pooling settings achieves significantly lower memory consumption, particularly in the initial layers where peak memory usage occurs.
 For DDPM, which utilizes a U-Net architecture with both down-sampling and up-sampling layers, our method reduces memory consumption in both the initial layers and the final layers where peak memory usage occurs.



\begin{table*}[t]
\small
\caption{\textbf{Edge Device Deployment.} We analyze the theoretical Peak Memory (T-PkMem), and measure actual GPU Peak Memory (A-PkMem), Model Size (MS), Maximum Power (MxP), and Latency for models on an NVIDIA Jetson TX2 device. For theoretical Peak Memory analysis, the batch size is assumed to be 1. For each actual measurement, the batch size (BS) is reported in the table and set to the maximum load allowed by the device, which is restricted by the teacher. The $1^{st}$-$9^{th}$ rows are for models we use in STL10 classification, and the $10^{th}$-$18^{th}$ row are for models we use in ImageNet classification, while the last 3 rows are for models we use in image generation task.
\rebuttal{The overheads of RED modules on top of student networks is minimal in particular for cases with more aggressive pooling, e.g., distillation of MbNetV2 with $\times 8$ pooling.}
}
\label{tab: edge device deployment}
\begin{center}
\setlength{\tabcolsep}{1.5mm}{
\begin{threeparttable}
\small
\begin{tabular}{l|c|c|c|c|c|c}
\hline
\hline
 
 \multirow{2}{*}{Model} & Theoretical Analysis & \multicolumn{5}{c}{Actual Measurement} \\
 \cline{2-7}
 & T-PkMem (MB) & BS & A-PkMem (GB) & MS (MB) & MxP (mW) & Latency (ms) \\
\hline

\hline
 T: MbNetV2 & 7.50 & \multirow{3}{*}{20} & 2.30 & \textbf{13.50} & 1830 & 221.53 $\pm$ 2.34 \\
 S: MbNetV2$\times 8$ & \textbf{0.51} & & \textbf{0.78} & \textbf{13.50} & \textbf{458} & \textbf{49.51 $\pm$ 1.98} \\
  \textbf{S w/ RED (ours)} & \textbf{0.51} & & \rebuttal{\textbf{0.78}} & 14.25 & \textbf{534} & \textbf{55.43 $\pm$ 1.80} \\
  
\hline
 T: MbNetV3-Small & 1.41 & \multirow{3}{*}{50} & 2.30 & \textbf{9.75} & 1526 & 195.63 $\pm$ 1.09 \\
 S: MbNetV3-Small$\times 4$ & \textbf{0.31} & & \textbf{1.10} & \textbf{9.75} & \textbf{611} & \textbf{69.76 $\pm$ 1.55} \\
 \textbf{S w/ RED (ours)} & \textbf{0.31} & & \textbf{1.10} & 10.18 & \textbf{687} & \textbf{83.74 $\pm$ 1.40} \\
 
\hline
 T: ResNext18 & 5.00 & \multirow{3}{*}{50} & 2.40 & \textbf{21.47} & 3276 & 448.38 $\pm$ 6.97 \\
 S: ResNext18$\times 4$ & \textbf{0.75} & & \textbf{1.10} & \textbf{21.47} & \textbf{1450} & \textbf{173.23 $\pm$ 1.92} \\
 \textbf{S w/ RED (ours)} & \textbf{0.75} & & \rebuttal{\textbf{1.10}} & 25.56 & \textbf{2058} & \textbf{275.02 $\pm$ 3.25} \\
\hline

\hline


T: ResNet18 & 3.83 & \multirow{3}{*}{50} & 2.70 & \textbf{44.63} & 3273 & 395.42 $\pm$ 4.21 \\
S: ResNet18$\times4$ & \textbf{0.77} & & \textbf{1.20} & \textbf{44.63} & \textbf{1297} & \textbf{151.59 $\pm$ 1.50} \\
\textbf{S w/ RED (ours)} & \textbf{0.77} & & \rebuttal{\textbf{1.20}} & 48.72 & \textbf{1830} & \textbf{208.43 $\pm$ 1.53} \\
\hline
T: ResNet50 & 9.19 & \multirow{3}{*}{10} & 2.00 & \textbf{97.70} & 2591 & 293.81 $\pm$ 1.09 \\
S: ResNet50$\times4$ & \textbf{2.30} & & \textbf{1.00} & \textbf{97.70} & \textbf{1068} & \textbf{125.49 $\pm$ 1.17} \\
\textbf{S w/ RED (ours)} & \textbf{2.30} & & \rebuttal{\textbf{1.00}} & 160.44 & \textbf{2136} & \textbf{246.68 $\pm$ 1.73} \\
\hline
T: ResNet152 & 9.19 & \multirow{3}{*}{8} & 2.80 & 230.20 & 3880 & 612.03 $\pm$ 3.55 \\
S: MobileNetV2$\times2$ & \textbf{1.15} & & \textbf{0.71} & \textbf{16.23} & \textbf{305} & \textbf{23.98 $\pm$ 0.84} \\
\textbf{S w/ RED (ours)} & \rebuttal{\textbf{1.15}} & & \rebuttal{\textbf{0.71}} & \textbf{32.73} & \textbf{611} & \textbf{72.65 $\pm$ 1.01} \\
\hline

\hline
T: UNet w/ DDPM & 3.50 & \multirow{3}{*}{30} & 2.90 & \textbf{133.09} & 4257 & 684.31 $\pm$ 5.39 \\
S: UNet$\times2$ w/ DDPM & \textbf{0.88} & & \textbf{1.30} & \textbf{133.09} & \textbf{1983} & \textbf{249.92 $\pm$ 3.76} \\
\textbf{S w/ RED (ours)} & \textbf{0.88} & & \rebuttal{\textbf{1.30}} & 149.92 & \textbf{2362} & \textbf{304.97 $\pm$ 1.87} \\

\hline
\hline
\end{tabular}
\end{threeparttable}
}
\end{center}
\end{table*}

 \paragraph{Edge Device Deployment}  
 In Table~\ref{tab: edge device deployment}, we measure theoretical peak memory consumption, actual peak memory consumption, and other efficiency related metrics like model size, maximum power, and latency, for all the models we use in STL10 \& ImageNet classification and image generation tasks.
 Specifically, the $1^{st}$-$9^{th}$ rows are for models we use in STL10 classification, and the $10^{th}$-$18^{th}$ row are for models we use in ImageNet classification, while the last 3 rows are for models we use in image generation task.
 We estimate the theoretical peak memory by summing the size of the input \& output allocation for each operation and assume a batch size of 1 for theoretical estimation to emulate most inference use-cases, similar to~(\cite{mcunetv2,chowdhery2019visual}).
 Our method achieves $3.9\times{\sim}14.7\times$ reduction in theoretical peak memory for image classification and $4\times$ reduction in theoretical peak memory for image generation, as shown in Table~\ref{tab: edge device deployment}. 
 In addition to the theoretical peak memory, we also measure the actual peak memory consumed on an NVIDIA Jetson TX2 device.
 Moreover, we measure the maximum GPU power (to ensure we satisfy the power budget of edge devices), model size, and latency incurred by the baseline teacher, student, and our RED models on the same edge GPU. 
 Our models yield similar peak memory as the student models (due to similar levels of aggressive striding) and ${\sim}2{\times}{\sim}{3.2}{\times}$ lower peak memory for image classification and ${\sim}2\times$ lower peak memory for image generation tasks compared to the teacher models. Note that our theoretical and measured peak memory reduction factors are different due to varying device setups and buffer allocations. 
 However, our models incur worse latency and power compared to the student models due to the additional RED blocks and improved latency compared to the teacher models.


\subsection{Ablation Study}
\label{sec: ablation study}



\begin{table}
\small
\caption{Module Discussion on STL10 dataset with MbNetV3-Small$\times 4$ with $\alpha{=}50$, and cosine distance for the RED loss. LM denotes Logit Module, and RE denotes Residual Encoder. ks denotes the kernel size of the convolution layer in RE.}
\label{tab: module discussion}
\begin{center}
\setlength{\tabcolsep}{1mm}{
\small
\begin{tabular}{l|c|c|c|c|c}
\hline
\hline
 \multirow{2}{*}{Method} & \multicolumn{4}{c|}{RED block} & \multirow{2}{*}{Top1 Acc. (\%)} \\
 \cline{2-5}
  & LM & RE & Shortcut & \rebuttal{ks} & \\
\hline 
 w/o LM & - & $\checkmark$ & $\checkmark$ & \rebuttal{3} & 76.66 \\
 w/o RE &  $\checkmark$ & - & $\checkmark$ & \rebuttal{-} & 72.30 \\
 w/o Shortcut & $\checkmark$ & $\checkmark$ & - & \rebuttal{3} & 52.42 \\
 w/o RED block &  - & - & - & \rebuttal{-} & 69.31 \\
 \rebuttal{RE w/ ks=1} & $\checkmark$ & $\checkmark$ & $\checkmark$ & \rebuttal{1} & \rebuttal{73.86} \\
 \rebuttal{RE w/ ks=5} & $\checkmark$ & $\checkmark$ & $\checkmark$ & \rebuttal{5} & \rebuttal{76.26} \\
\hline
 \textbf{RED (ours)} & $\checkmark$ & $\checkmark$ & $\checkmark$ & \rebuttal{3} & \textbf{77.31} \\
\hline
\hline
\end{tabular}
}
\end{center}
\vspace{-4mm}
\end{table}

 \paragraph{Module Discussion} As shown in Table~\ref{tab: module discussion}, we conduct the ablation study on STL10 dataset with the MobileNetV3-Small backbone by removing the logit module, residual encoder, shortcut, and the whole RED block, respectively.
 The performance of our proposed model degrades slightly without the logit module but seriously degrades without the residual encoder, which illustrates that the residual encoder plays a more significant role in the RED block.
 Without RED blocks, the student model has poor performance when only applying RED loss on the student activation maps and teacher activation maps, which illustrates the necessity to integrate RED blocks into the student model.
 Naively stacking the logit module and residual encoder, i.e., without Shortcut, performs even worse than without RED blocks, illustrating the effectiveness of our designed shortcuts in the logit module and residual encoder.
 \rebuttal{We also conduct the ablation study on the kernel size of the residual module of our RED block.
 It shows that the proposed method achieves the best performance with kernel size 3.}

\paragraph{Distillation Strategy}
 We conduct the ablation study to compare different strategies during distillation, including the feature alignment strategy and pooling stride of the initial layer.
 Specifically, we train the ResNext18 network on the STL10 dataset with three different strategies while enlarging the stride of the first pooling layer to $\times2$, $\times4$, and $\times8$.
 The results are shown in Table~\ref{tab: pooling strategy discussion}.
 `RED w/o Distillation' is the baseline of training an aggressive pooled ResNext18 without distillation.
 `RED w/ Stage-Align' applies the traditional stage-based feature alignment between the student and the teacher activations for distillation.
 `RED w/ Pooling-Align' is the proposed pooling-based feature alignment strategy.
 Compared to traditional stage-based alignment, the proposed pooling-based alignment performs much better for the aggressive pooling strategy.

\begin{table}[t]
\caption{Ablation study of distillation strategies on STL10 dataset with ResNext18.}
\vspace{-2mm}
\label{tab: pooling strategy discussion}
\begin{center}
\setlength{\tabcolsep}{1mm}{
\begin{threeparttable}
\begin{tabular}{l|c|c|c}
\hline
\hline
 Backbone: ResNext18 & \multicolumn{3}{c}{Top1 Acc. (\%)} \\
\hline
 $1^{st}$ Pooling Stride & $\times{2}$ & $\times{4}$ & $\times{8}$ \\

\hline 
 RED w/o Distillation & 81.43 & 79.07 & 73.16 \\
 RED w/ Stage-Align & 80.59 & 81.99 & 74.00 \\
 \textbf{RED w/ Pooling-Align (ours)} & \textbf{85.51} & \textbf{84.80} & \textbf{77.48} \\ 

\hline
\hline
\end{tabular}

\end{threeparttable}
}
\end{center}
\vspace{-3mm}
\end{table}

\begin{wraptable}{r}{9.4cm}
\vspace{-5mm}
\caption{Loss Function Discussion on STL10 dataset with ResNext18$\times 4$.
\rebuttal{We conduct ablation study on $\alpha$ and distance measurement for $\mathcal{L}_{RED}$.
We also compare our method with the student integrated with the proposed RED block but distilled only by $\mathcal{L}_{KD}$ and distilled by both our $\mathcal{L}_{RED}$ and $\mathcal{L}_{KD}$.}
}
\label{tab: loss function discussion}
\begin{center}
\setlength{\tabcolsep}{1mm}{
\begin{threeparttable}
\begin{tabular}{l|c|c|c|c}
\hline
\hline
 \multirow{2}{*}{Method} & \multicolumn{2}{c|}{$\mathcal{L}_{RED}$} & \multirow{2}{*}{$\mathcal{L}_{KD}$} & \multirow{2}{*}{Top1 Acc. (\%)} \\
 \cline{2-3}
   & $\alpha$ & Distance & & \\

\hline 

 \multirow{8}{*}{\textbf{RED (ours)}} & 0 & - & - & 82.26 \\
  & 50 & Euclidean & - & 82.66 \\
  & 1 & Cosine & - & 82.04 \\
  & \rebuttal{5} & Cosine & - & \rebuttal{82.86} \\
  & \rebuttal{10} & Cosine & - & \rebuttal{83.41} \\
  & 50 & Cosine & - & 84.80 \\
  & \rebuttal{100} & Cosine & - & \rebuttal{\textbf{85.10}} \\
  & \rebuttal{200} & Cosine & - & \rebuttal{84.95} \\
 \hline
 \textbf{RED$_{w/o \mathcal{L}_{RED}}$+KD} & 0 & - & $\checkmark$ & 81.94 \\
 \textbf{RED (ours)+KD} & 50 & Cosine & $\checkmark$ & 84.82 \\

\hline
\hline
\end{tabular}

\end{threeparttable}
}
\end{center}
\vspace{-4mm}
\end{wraptable}

 \paragraph{Loss Function}  
 We also conduct the ablation study for the proposed RED loss function, an equally important portion of distillation.
 The results are based on ResNext18 backbone and STL10 dataset, as shown in Table~\ref{tab: loss function discussion}.
 Without the RED loss, i.e. setting $\alpha$ to $0$ and then the loss function $\mathcal{L} = \mathcal{L}_{task}$, the student model is just integrated with RED blocks but without distillation.
 In this circumstance, the poor performance of the student model illustrates the need to apply the loss of RED.
 The proposed method performs worse when applying the Euclidean distance instead of the cosine distance to calculate the RED loss.
 Averaging along the channel dimension with different $C_T$ and $C_S$ in Equation~\ref{eq:RED block loss} might mitigate the absolute difference in Euclidean space, while cosine distance measures the angle between two vectors, preserving their relative difference.
 By adjusting $\alpha$ from $200$ to $1$, the performance degradation shows that the proposed method is sensitive to the hyper-parameter $\alpha$ in Equation~\ref{eq: final loss}.
 Besides, we evaluate the proposed method's performance when incorporated with other knowledge distillation methods, such as KD~(\cite{KD2015}).
 Specifically, we first evaluate the student model integrated with the proposed RED blocks by applying only KD loss, which is denoted as `RED$_{w/o \mathcal{L}_{RED}}$+KD'.
 Then we apply both RED loss and KD loss into the same student model, denoted as `RED (ours)+KD'.
 Performance improvement illustrates the effectiveness of the proposed RED loss.

\begin{wraptable}{r}{5.8cm}
\vspace{-5.2mm}
\caption{Distillation w/o Aggressive Pooling Setting.}
\vspace{-2mm}
\label{tab: comparison w/o aggressive pooling}
\begin{center}
\setlength{\tabcolsep}{1mm}{
\begin{threeparttable}
\begin{tabular}{l|c}
\hline
\hline
Method & Top1 Acc. (\%) \\
\hline
T: ResNext50 & 83.54 \\
S: MbNetV3-Small & 72.62 \\
\hline
KD & 77.06 \\
FitNet & 76.35 \\
AT & 76.76 \\
VID & 74.83 \\
RKD & 73.96 \\
AB & 77.50 \\
NST & 73.78 \\
\hline
RED (ours) & \textbf{79.71} \\

\hline
\hline
\end{tabular}

\end{threeparttable}
}
\end{center}
\vspace{-16mm}
\end{wraptable}

 \paragraph{Distillation w/o Aggressive Pooling} In addition to aggressive pooling, we conduct experiments to compare several distillation methods from the teacher network ResNext50 to the student network MobileNetV3-Small without aggressive pooling on STL10 dataset.
 As shown in Table~\ref{tab: comparison w/o aggressive pooling}, our method still achieves the best performance, illustrating the generalization ability of the proposed framework.





\vspace{-6mm}
\rebuttal{
 \paragraph{Comparison to Non-distillation Method} We compare our methods to quantization-aware-training (QAT) method by using the pytorch native architecture optimization toolkit TorchAO~(\cite{torchao}) and a 6 bits QAT method, INQ~(\cite{inq}).
 We also deploy Intel Neural Compressor (INC)~(\cite{inc}) to conduct int8 fully precision and int8/fp32 mixed precision post-training-quantization (PTQ) methods.
 As shown in Table~\ref{tab: comparison w/ non-distillation method}, the ResNext18 quantized by INC to int8 achieves 20.12\% accuracy and 1.25MB peak memory.
 The ResNext18 quantized by INC to mixed precision achieves 84.18\% accuracy and 4.19 MB peak memory.
 The ResNext18 quantized by INQ to 6 bits achieves 84.07\% accuracy and 5.00 MB peak memory. 
 Our distillation method achieves the optimal trade-off between accuracy and peak memory, with 84.80\% accuracy and 0.75 MB peak memory.
 Due to the extra RED blocks, our method possesses the higher model size of 25.557 MB.
 However, our method is orthogonal to these quantization methods and thus can be applied with the quantization method to further optimize the student model’s overhead.
 As shown in the last row of Table~\ref{tab: comparison w/ non-distillation method}, the student model ResNext18×4 distilled by our method and quantized by INC to mixed
precision achieves 84.78\% accuracy, 0.44 MB peak memory, and 5.621 MB model size.
}

\begin{table}[h]
\caption{Comparison w/ Non-distillation Method.}
\label{tab: comparison w/ non-distillation method}
\begin{center}
\setlength{\tabcolsep}{1mm}{
\begin{tabular}{l|c|c|c}
\hline
\hline
Method & Top1 Acc.(\%) & T-PkMem (MB) & Model Size (MB) \\
\hline
T: ResNext18 - fp32 & \textbf{85.12} & 5.00 & 21.47 \\

\quad\quad \rebuttal{w/ TorchAO - dynamic, QAT} & 83.37 & 5.00 & 19.513 \\
\quad\quad \rebuttal{w/ INC - int8, PTQ} & \rebuttal{20.12} & \rebuttal{1.25} & \rebuttal{4.960} \\
\quad\quad \rebuttal{w/ INC - mixed prec., PTQ} & \rebuttal{84.18} & \rebuttal{4.19} & \rebuttal{5.015} \\
\quad\quad \rebuttal{w/ INQ - 6 bits, QAT} & \rebuttal{84.07} & \rebuttal{5.00} & \rebuttal{\textbf{4.026}} \\
\hline
S: ResNext18$\times 4$ - fp32 & 79.07 & 0.75 & 21.47 \\
\quad\quad \rebuttal{w/ INC - mixed prec., PTQ} & \rebuttal{78.50} & \rebuttal{\textbf{0.44}} & \rebuttal{5.015} \\
\hline
S + \textbf{RED (ours)} - fp32 & \textbf{84.80} & 0.75 & 25.557 \\
\quad\quad \rebuttal{w/ INC - mixed prec., PTQ} & \rebuttal{84.78} & \rebuttal{\textbf{0.44}} & \rebuttal{5.621} \\

\hline
\hline
\end{tabular}
}
\end{center}
\end{table}

\section{Conclusions \rebuttal{and Future Work}}
\label{conclusion}

We propose ReDistill, a novel residual encoded distillation method to reduce the peak memory of convolutional neural networks during inference. Our method enables the deployment of these networks in edge devices, such as micro-controllers with tight memory budgets, while accommodating high-resolution images necessary for intricate vision tasks. The reduced peak memory can also enable these networks to be implemented with recently proposed in-sensor computing systems~(\cite{aps_p2m,datta2023in-sensor1}), thereby significantly reducing the bandwidth between the image sensor and the back-end processing unit. 
Our method is based on a teacher-student distillation framework, where the student network using aggressive pooling with reduced peak memory is distilled from the teacher network. For image classification, our method outperforms existing response-based and feature-based distillation methods in terms of accuracy-memory trade-off. For diffusion-based image generation, our method significantly reduces the peak memory of the denoising network with slight degradation in the fidelity and diversity of the generated images.
\rebuttal{While we focus on distillation of CNNs in this work, we leave as future work peak memory reduction of vision transformer via distillation for image classification~(\cite{tinyvit2022}) and image generation~(\cite{dit2023}).}


\paragraph{Acknowledgement} This work is supported by Department of Defense under funding award W911NF-241-0295, National Science Foundation Career award \#2341039, and Google Cloud research credits program.

\bibliography{main}

\begin{thebibliography}{59}
\providecommand{\natexlab}[1]{#1}
\providecommand{\url}[1]{\texttt{#1}}
\expandafter\ifx\csname urlstyle\endcsname\relax
  \providecommand{\doi}[1]{doi: #1}\else
  \providecommand{\doi}{doi: \begingroup \urlstyle{rm}\Url}\fi

\bibitem[Adriana et~al.(2015)Adriana, Nicolas, Ebrahimi, Antoine, Carlo, and Yoshua]{FITNET2015}
Romero Adriana, Ballas Nicolas, K~Samira Ebrahimi, Chassang Antoine, Gatta Carlo, and Bengio Yoshua.
\newblock Fitnets: Hints for thin deep nets.
\newblock \emph{In Proceedings of the International Conference on Learning Representations}, 2\penalty0 (3):\penalty0 1, 2015.

\bibitem[Ahn et~al.(2019)Ahn, Hu, Damianou, Lawrence, and Dai]{VID2019}
Sungsoo Ahn, Shell~Xu Hu, Andreas Damianou, Neil~D Lawrence, and Zhenwen Dai.
\newblock Variational information distillation for knowledge transfer.
\newblock In \emph{Proceedings of the IEEE/CVF conference on computer vision and pattern recognition}, pp.\  9163--9171, 2019.

\bibitem[Aojun~Zhou et~al.(2017)Aojun~Zhou, Yiwen~Guo, and Chen]{inq}
Anbang~Yao Aojun~Zhou, Lin~Xu Yiwen~Guo, and Yurong Chen.
\newblock Incremental network quantization: Towards lossless cnns with low-precision weights.
\newblock In \emph{International Conference on Learning Representations, ICLR2017}, 2017.

\bibitem[Bao et~al.(2023)Bao, Nie, Xue, Cao, Li, Su, and Zhu]{uvit2023}
Fan Bao, Shen Nie, Kaiwen Xue, Yue Cao, Chongxuan Li, Hang Su, and Jun Zhu.
\newblock All are worth words: A vit backbone for diffusion models.
\newblock In \emph{Proceedings of the IEEE/CVF Conference on Computer Vision and Pattern Recognition}, pp.\  22669--22679, 2023.

\bibitem[Chen et~al.(2023)Chen, Datta, Kundu, and Beerel]{chen2023self}
Fang Chen, Gourav Datta, Souvik Kundu, and Peter~A Beerel.
\newblock Self-attentive pooling for efficient deep learning.
\newblock In \emph{Proceedings of the IEEE/CVF Winter Conference on Applications of Computer Vision}, pp.\  3974--3983, 2023.

\bibitem[Chen et~al.(2021)Chen, Liu, Zhao, and Jia]{ReviewKD2021}
Pengguang Chen, Shu Liu, Hengshuang Zhao, and Jiaya Jia.
\newblock Distilling knowledge via knowledge review.
\newblock In \emph{Proceedings of the IEEE/CVF Conference on Computer Vision and Pattern Recognition}, pp.\  5008--5017, 2021.

\bibitem[Cheng et~al.(2023)Cheng, Cai, Lv, and Shen]{inc}
Wenhua Cheng, Yiyang Cai, Kaokao Lv, and Haihao Shen.
\newblock Teq: Trainable equivalent transformation for quantization of llms, 2023.
\newblock URL \url{https://arxiv.org/abs/2310.10944}.

\bibitem[Chowdhery et~al.(2019)Chowdhery, Warden, Shlens, Howard, and Rhodes]{chowdhery2019visual}
Aakanksha Chowdhery, Pete Warden, Jonathon Shlens, Andrew Howard, and Rocky Rhodes.
\newblock Visual wake words dataset.
\newblock \emph{arXiv preprint arXiv:1906.05721}, 2019.

\bibitem[Coates et~al.(2011)Coates, Ng, and Lee]{coates2011analysis}
Adam Coates, Andrew Ng, and Honglak Lee.
\newblock An analysis of single-layer networks in unsupervised feature learning.
\newblock In \emph{Proceedings of the fourteenth international conference on artificial intelligence and statistics}, pp.\  215--223. JMLR Workshop and Conference Proceedings, 2011.

\bibitem[Creswell et~al.(2018)Creswell, White, Dumoulin, Arulkumaran, Sengupta, and Bharath]{creswell2018generative}
Antonia Creswell, Tom White, Vincent Dumoulin, Kai Arulkumaran, Biswa Sengupta, and Anil~A Bharath.
\newblock Generative adversarial networks: An overview.
\newblock \emph{IEEE signal processing magazine}, 35\penalty0 (1):\penalty0 53--65, 2018.

\bibitem[Datta et~al.(2022)Datta, Kundu, Yin, Lakkireddy, Mathai, Jacob, Beerel, and Jaiswal]{aps_p2m}
Gourav Datta, Souvik Kundu, Zihan Yin, Ravi~Teja Lakkireddy, Joe Mathai, Ajey~P Jacob, Peter~A Beerel, and Akhilesh~R Jaiswal.
\newblock A processing-in-pixel-in-memory paradigm for resource-constrained tinyml applications.
\newblock \emph{Scientific Reports}, 12\penalty0 (1):\penalty0 14396, 2022.

\bibitem[Datta et~al.(2023)Datta, Liu, Kaiser, Kundu, Mathai, Yin, Jacob, Jaiswal, and Beerel]{datta2023in-sensor1}
Gourav Datta, Zeyu Liu, Md~Abdullah-Al Kaiser, Souvik Kundu, Joe Mathai, Zihan Yin, Ajey~P. Jacob, Akhilesh~R. Jaiswal, and Peter~A. Beerel.
\newblock In-sensor \& neuromorphic computing are all you need for energy efficient computer vision.
\newblock In \emph{ICASSP 2023 - 2023 IEEE International Conference on Acoustics, Speech and Signal Processing (ICASSP)}, pp.\  1--5, 2023.
\newblock \doi{10.1109/ICASSP49357.2023.10094902}.

\bibitem[Han et~al.(2016{\natexlab{a}})Han, Liu, Mao, Pu, Pedram, Horowitz, and Dally]{han2016eie}
Song Han, Xingyu Liu, Huizi Mao, Jing Pu, Ardavan Pedram, Mark~A Horowitz, and William~J Dally.
\newblock Eie: Efficient inference engine on compressed deep neural network.
\newblock \emph{ACM SIGARCH Computer Architecture News}, 44\penalty0 (3):\penalty0 243--254, 2016{\natexlab{a}}.

\bibitem[Han et~al.(2016{\natexlab{b}})Han, Mao, and Dally]{han2016deep}
Song Han, Huizi Mao, and William~J Dally.
\newblock Deep compression: Compressing deep neural networks with pruning, trained quantization and huffman coding.
\newblock \emph{International Conference on Learning Representations}, 2016{\natexlab{b}}.

\bibitem[Hao et~al.(2023)Hao, Guo, Han, Tang, Hu, Wang, and Xu]{ofakd}
Zhiwei Hao, Jianyuan Guo, Kai Han, Yehui Tang, Han Hu, Yunhe Wang, and Chang Xu.
\newblock One-for-all: Bridge the gap between heterogeneous architectures in knowledge distillation.
\newblock In \emph{Advances in Neural Information Processing Systems}, 2023.

\bibitem[He et~al.(2016)He, Zhang, Ren, and Sun]{he2016deep}
Kaiming He, Xiangyu Zhang, Shaoqing Ren, and Jian Sun.
\newblock Deep residual learning for image recognition.
\newblock In \emph{Proceedings of the IEEE conference on computer vision and pattern recognition}, pp.\  770--778, 2016.

\bibitem[Heo et~al.(2019)Heo, Lee, Yun, and Choi]{Abound2018}
Byeongho Heo, Minsik Lee, Sangdoo Yun, and Jin~Young Choi.
\newblock Knowledge transfer via distillation of activation boundaries formed by hidden neurons.
\newblock In \emph{Proceedings of the AAAI conference on artificial intelligence}, volume~33, pp.\  3779--3787, 2019.

\bibitem[Heusel et~al.(2017)Heusel, Ramsauer, Unterthiner, Nessler, and Hochreiter]{heusel2018gans}
Martin Heusel, Hubert Ramsauer, Thomas Unterthiner, Bernhard Nessler, and Sepp Hochreiter.
\newblock Gans trained by a two time-scale update rule converge to a local nash equilibrium.
\newblock \emph{Advances in neural information processing systems}, 30, 2017.

\bibitem[Hinton et~al.(2015)Hinton, Vinyals, and Dean]{KD2015}
Geoffrey Hinton, Oriol Vinyals, and Jeff Dean.
\newblock Distilling the knowledge in a neural network.
\newblock \emph{arXiv preprint arXiv:1503.02531}, 2015.

\bibitem[Ho et~al.(2020)Ho, Jain, and Abbeel]{ddpm2020}
Jonathan Ho, Ajay Jain, and Pieter Abbeel.
\newblock Denoising diffusion probabilistic models.
\newblock \emph{Advances in neural information processing systems}, 33:\penalty0 6840--6851, 2020.

\bibitem[Howard et~al.(2019)Howard, Sandler, Chu, Chen, Chen, Tan, Wang, Zhu, Pang, Vasudevan, et~al.]{mobilenetv3}
Andrew Howard, Mark Sandler, Grace Chu, Liang-Chieh Chen, Bo~Chen, Mingxing Tan, Weijun Wang, Yukun Zhu, Ruoming Pang, Vijay Vasudevan, et~al.
\newblock Searching for mobilenetv3.
\newblock In \emph{Proceedings of the IEEE/CVF international conference on computer vision}, pp.\  1314--1324, 2019.

\bibitem[Huang \& Wang(2017)Huang and Wang]{NST2017}
Zehao Huang and Naiyan Wang.
\newblock Like what you like: Knowledge distill via neuron selectivity transfer.
\newblock \emph{arXiv preprint arXiv:1707.01219}, 2017.

\bibitem[Hubara et~al.(2016)Hubara, Courbariaux, Soudry, El-Yaniv, and Bengio]{hubara2016binarized}
Itay Hubara, Matthieu Courbariaux, Daniel Soudry, Ran El-Yaniv, and Yoshua Bengio.
\newblock Binarized neural networks.
\newblock \emph{Advances in neural information processing systems}, 29, 2016.

\bibitem[Hunter(1986)]{hunter1986exponentially}
J~Stuart Hunter.
\newblock The exponentially weighted moving average.
\newblock \emph{Journal of quality technology}, 18\penalty0 (4):\penalty0 203--210, 1986.

\bibitem[Iandola et~al.(2017)Iandola, Han, Moskewicz, Ashraf, Dally, and Keutzer]{iandola2017squeezenet}
Forrest~N Iandola, Song Han, Matthew~W Moskewicz, Khalid Ashraf, William~J Dally, and Kurt Keutzer.
\newblock Squeezenet: Alexnet-level accuracy with 50x fewer parameters and< 0.5 mb model size.
\newblock \emph{International Conference on Learning Representations}, 2017.

\bibitem[Jin et~al.(2023)Jin, Wang, and Lin]{MLLD2023}
Ying Jin, Jiaqi Wang, and Dahua Lin.
\newblock Multi-level logit distillation.
\newblock In \emph{Proceedings of the IEEE/CVF Conference on Computer Vision and Pattern Recognition}, pp.\  24276--24285, 2023.

\bibitem[Kim et~al.(2018)Kim, Park, and Kwak]{Factor2020}
Jangho Kim, SeongUk Park, and Nojun Kwak.
\newblock Paraphrasing complex network: Network compression via factor transfer.
\newblock \emph{Advances in neural information processing systems}, 31, 2018.

\bibitem[Krizhevsky et~al.(2009)Krizhevsky, Hinton, et~al.]{alex2009learning}
Alex Krizhevsky, Geoffrey Hinton, et~al.
\newblock Learning multiple layers of features from tiny images.
\newblock \emph{Technical Report, University of Toronto}, 2009.

\bibitem[Langer et~al.(2020)Langer, He, Rahayu, and Xue]{langer2020distributed}
Matthias Langer, Zhen He, Wenny Rahayu, and Yanbo Xue.
\newblock Distributed training of deep learning models: A taxonomic perspective.
\newblock \emph{IEEE Transactions on Parallel and Distributed Systems}, 31\penalty0 (12):\penalty0 2802--2818, 2020.

\bibitem[Lee et~al.(2018)Lee, Kim, and Song]{KDSVD2018}
Seung~Hyun Lee, Dae~Ha Kim, and Byung~Cheol Song.
\newblock Self-supervised knowledge distillation using singular value decomposition.
\newblock In \emph{Proceedings of the European conference on computer vision (ECCV)}, pp.\  335--350, 2018.

\bibitem[Li et~al.(2022)Li, Lin, Ding, Lin, Zhuang, Huang, Ding, and Cao]{KCD2022}
Chenxin Li, Mingbao Lin, Zhiyuan Ding, Nie Lin, Yihong Zhuang, Yue Huang, Xinghao Ding, and Liujuan Cao.
\newblock Knowledge condensation distillation.
\newblock In \emph{European Conference on Computer Vision}, pp.\  19--35. Springer, 2022.

\bibitem[Li et~al.(2023)Li, Li, Zheng, Wu, Xiao, Wang, Zheng, Pan, Chao, and Ji]{li2023autodiffusion}
Lijiang Li, Huixia Li, Xiawu Zheng, Jie Wu, Xuefeng Xiao, Rui Wang, Min Zheng, Xin Pan, Fei Chao, and Rongrong Ji.
\newblock Autodiffusion: Training-free optimization of time steps and architectures for automated diffusion model acceleration.
\newblock In \emph{Proceedings of the IEEE/CVF International Conference on Computer Vision}, pp.\  7105--7114, 2023.

\bibitem[Lin et~al.(2020)Lin, Chen, Lin, Gan, Han, et~al.]{mcunet}
Ji~Lin, Wei-Ming Chen, Yujun Lin, Chuang Gan, Song Han, et~al.
\newblock Mcunet: Tiny deep learning on iot devices.
\newblock \emph{Advances in Neural Information Processing Systems}, 33:\penalty0 11711--11722, 2020.

\bibitem[Lin et~al.(2021)Lin, Chen, Cai, Gan, and Han]{mcunetv2}
Ji~Lin, Wei-Ming Chen, Han Cai, Chuang Gan, and Song Han.
\newblock Mcunetv2: Memory-efficient patch-based inference for tiny deep learning.
\newblock \emph{arXiv preprint arXiv:2110.15352}, 2021.

\bibitem[Lin et~al.(2022)Lin, Zhu, Chen, Wang, Gan, and Han]{mcunetv3}
Ji~Lin, Ligeng Zhu, Wei-Ming Chen, Wei-Chen Wang, Chuang Gan, and Song Han.
\newblock On-device training under 256kb memory.
\newblock \emph{Advances in Neural Information Processing Systems}, 35:\penalty0 22941--22954, 2022.

\bibitem[Liu et~al.(2015)Liu, Luo, Wang, and Tang]{liu2015faceattributes}
Ziwei Liu, Ping Luo, Xiaogang Wang, and Xiaoou Tang.
\newblock Deep learning face attributes in the wild.
\newblock In \emph{Proceedings of International Conference on Computer Vision (ICCV)}, December 2015.

\bibitem[Long et~al.(2015)Long, Shelhamer, and Darrell]{long2015fully}
Jonathan Long, Evan Shelhamer, and Trevor Darrell.
\newblock Fully convolutional networks for semantic segmentation.
\newblock In \emph{Proceedings of the IEEE conference on computer vision and pattern recognition}, pp.\  3431--3440, 2015.

\bibitem[Lu et~al.(2024)Lu, Guan, Yang, Zhao, Gong, and Xu]{lu2024entropy}
Yiheng Lu, Ziyu Guan, Yaming Yang, Wei Zhao, Maoguo Gong, and Cai Xu.
\newblock Entropy induced pruning framework for convolutional neural networks.
\newblock In \emph{Proceedings of the AAAI Conference on Artificial Intelligence}, volume~38, pp.\  3918--3926, 2024.

\bibitem[Molchanov et~al.(2017)Molchanov, Tyree, Karras, Aila, and Kautz]{molchanov2017pruning}
Pavlo Molchanov, Stephen Tyree, Tero Karras, Timo Aila, and Jan Kautz.
\newblock Pruning convolutional neural networks for resource efficient inference.
\newblock \emph{International Conference on Learning Representations}, 2017.

\bibitem[Park et~al.(2019)Park, Kim, Lu, and Cho]{RKD2019}
Wonpyo Park, Dongju Kim, Yan Lu, and Minsu Cho.
\newblock Relational knowledge distillation.
\newblock In \emph{Proceedings of the IEEE/CVF conference on computer vision and pattern recognition}, pp.\  3967--3976, 2019.

\bibitem[Passalis \& Tefas(2018)Passalis and Tefas]{PKT2018}
Nikolaos Passalis and Anastasios Tefas.
\newblock Learning deep representations with probabilistic knowledge transfer.
\newblock In \emph{Proceedings of the European Conference on Computer Vision (ECCV)}, pp.\  268--284, 2018.

\bibitem[Peebles \& Xie(2023)Peebles and Xie]{dit2023}
William Peebles and Saining Xie.
\newblock Scalable diffusion models with transformers.
\newblock In \emph{Proceedings of the IEEE/CVF international conference on computer vision}, pp.\  4195--4205, 2023.

\bibitem[Redmon \& Farhadi(2018)Redmon and Farhadi]{redmon2018yolov3}
Joseph Redmon and Ali Farhadi.
\newblock Yolov3: An incremental improvement.
\newblock \emph{arXiv preprint arXiv:1804.02767}, 2018.

\bibitem[Ronneberger et~al.(2015)Ronneberger, Fischer, and Brox]{ronneberger2015u}
Olaf Ronneberger, Philipp Fischer, and Thomas Brox.
\newblock U-net: Convolutional networks for biomedical image segmentation.
\newblock In \emph{Medical image computing and computer-assisted intervention--MICCAI 2015: 18th international conference, Munich, Germany, October 5-9, 2015, proceedings, part III 18}, pp.\  234--241. Springer, 2015.

\bibitem[Russakovsky et~al.(2015)Russakovsky, Deng, Su, Krause, Satheesh, Ma, Huang, Karpathy, Khosla, Bernstein, Berg, and Fei-Fei]{ILSVRC15}
Olga Russakovsky, Jia Deng, Hao Su, Jonathan Krause, Sanjeev Satheesh, Sean Ma, Zhiheng Huang, Andrej Karpathy, Aditya Khosla, Michael Bernstein, Alexander~C. Berg, and Li~Fei-Fei.
\newblock {ImageNet Large Scale Visual Recognition Challenge}.
\newblock \emph{International Journal of Computer Vision (IJCV)}, 115\penalty0 (3):\penalty0 211--252, 2015.
\newblock \doi{10.1007/s11263-015-0816-y}.

\bibitem[Salimans et~al.(2016)Salimans, Goodfellow, Zaremba, Cheung, Radford, and Chen]{salimans2016improved}
Tim Salimans, Ian Goodfellow, Wojciech Zaremba, Vicki Cheung, Alec Radford, and Xi~Chen.
\newblock Improved techniques for training gans.
\newblock \emph{Advances in neural information processing systems}, 29, 2016.

\bibitem[Sandler et~al.(2018)Sandler, Howard, Zhu, Zhmoginov, and Chen]{mobilenetv2}
Mark Sandler, Andrew Howard, Menglong Zhu, Andrey Zhmoginov, and Liang-Chieh Chen.
\newblock Mobilenetv2: Inverted residuals and linear bottlenecks.
\newblock In \emph{Proceedings of the IEEE conference on computer vision and pattern recognition}, pp.\  4510--4520, 2018.

\bibitem[Sauer et~al.(2023)Sauer, Lorenz, Blattmann, and Rombach]{sauer2023adversarial}
Axel Sauer, Dominik Lorenz, Andreas Blattmann, and Robin Rombach.
\newblock Adversarial diffusion distillation.
\newblock \emph{arXiv preprint arXiv:2311.17042}, 2023.

\bibitem[Simonyan \& Zisserman(2014)Simonyan and Zisserman]{simonyan2014very}
Karen Simonyan and Andrew Zisserman.
\newblock Very deep convolutional networks for large-scale image recognition.
\newblock \emph{arXiv preprint arXiv:1409.1556}, 2014.

\bibitem[Song et~al.(2020)Song, Meng, and Ermon]{ddim2020}
Jiaming Song, Chenlin Meng, and Stefano Ermon.
\newblock Denoising diffusion implicit models.
\newblock \emph{arXiv preprint arXiv:2010.02502}, 2020.

\bibitem[Sun et~al.(2024)Sun, Ren, Li, Wang, and Cao]{lskd}
Shangquan Sun, Wenqi Ren, Jingzhi Li, Rui Wang, and Xiaochun Cao.
\newblock Logit standardization in knowledge distillation.
\newblock In \emph{Proceedings of the IEEE/CVF Conference on Computer Vision and Pattern Recognition}, pp.\  15731--15740, 2024.

\bibitem[Tian et~al.(2020)Tian, Krishnan, and Isola]{crd2020}
Yonglong Tian, Dilip Krishnan, and Phillip Isola.
\newblock Contrastive representation distillation.
\newblock In \emph{International Conference on Learning Representations}, 2020.

\bibitem[torchao maintainers \& contributors(2024)torchao maintainers and contributors]{torchao}
torchao maintainers and contributors.
\newblock torchao: Pytorch native quantization and sparsity for training and inference, October 2024.
\newblock URL \url{https://github.com/pytorch/torchao}.

\bibitem[Tung \& Mori(2019)Tung and Mori]{Similarity2019}
Frederick Tung and Greg Mori.
\newblock Similarity-preserving knowledge distillation.
\newblock In \emph{Proceedings of the IEEE/CVF international conference on computer vision}, pp.\  1365--1374, 2019.

\bibitem[Wu et~al.(2022)Wu, Zhang, Peng, Liu, Xiao, Fu, and Yuan]{tinyvit2022}
Kan Wu, Jinnian Zhang, Houwen Peng, Mengchen Liu, Bin Xiao, Jianlong Fu, and Lu~Yuan.
\newblock Tinyvit: Fast pretraining distillation for small vision transformers.
\newblock In \emph{European conference on computer vision}, pp.\  68--85. Springer, 2022.

\bibitem[Xie et~al.(2017)Xie, Girshick, Doll{\'a}r, Tu, and He]{resnext}
Saining Xie, Ross Girshick, Piotr Doll{\'a}r, Zhuowen Tu, and Kaiming He.
\newblock Aggregated residual transformations for deep neural networks.
\newblock In \emph{Proceedings of the IEEE conference on computer vision and pattern recognition}, pp.\  1492--1500, 2017.

\bibitem[Yin et~al.(2023)Yin, Gharbi, Zhang, Shechtman, Durand, Freeman, and Park]{yin2023one}
Tianwei Yin, Micha{\"e}l Gharbi, Richard Zhang, Eli Shechtman, Fredo Durand, William~T Freeman, and Taesung Park.
\newblock One-step diffusion with distribution matching distillation.
\newblock \emph{arXiv preprint arXiv:2311.18828}, 2023.

\bibitem[Zagoruyko \& Komodakis(2016)Zagoruyko and Komodakis]{AT2016}
Sergey Zagoruyko and Nikos Komodakis.
\newblock Paying more attention to attention: Improving the performance of convolutional neural networks via attention transfer.
\newblock \emph{arXiv preprint arXiv:1612.03928}, 2016.

\bibitem[Zhao et~al.(2022)Zhao, Cui, Song, Qiu, and Liang]{DKD2022}
Borui Zhao, Quan Cui, Renjie Song, Yiyu Qiu, and Jiajun Liang.
\newblock Decoupled knowledge distillation.
\newblock In \emph{Proceedings of the IEEE/CVF Conference on computer vision and pattern recognition}, pp.\  11953--11962, 2022.

\end{thebibliography}
\bibliographystyle{tmlr}

\end{document}